%% file: main.tex
\documentclass[10pt,twocolumn,letterpaper]{article}

\usepackage{cvpr}              

\usepackage[accsupp]{axessibility} 

\input{preamble}

\definecolor{cvprblue}{rgb}{0.21,0.49,0.74}
\usepackage[pagebackref,breaklinks,colorlinks,citecolor=cvprblue]{hyperref}

\usepackage{booktabs}
\usepackage{multirow}
\usepackage{graphicx}
\usepackage{amssymb}
\usepackage{xspace}
\usepackage{arydshln}
\usepackage{makecell}
\usepackage{marvosym}
\usepackage{color,colortbl}
\usepackage{tcolorbox}
\tcbuselibrary{skins}
\usepackage{array}
\usepackage{tabularx}
\tcbuselibrary{breakable}

\usepackage{bbding} 
\usepackage{enumitem}
\usepackage{svg}
\usepackage{subcaption}

\definecolor{myblue}{RGB}{71,135,237}

\newcommand\blfootnote[1]{%
  \begingroup
  \renewcommand\thefootnote{}\footnote{#1}%
  \addtocounter{footnote}{-1}%
  \endgroup
}

\def\paperID{14675} 
\def\confName{CVPR}
\def\confYear{2024}

\title{EgoThink: Evaluating First-Person Perspective Thinking Capability of Vision-Language Models}

\author{\textbf{Sijie Cheng}\textsuperscript{1,2,5 \ddag}\thanks{Equal contribution, \textsuperscript{\ddag} Project lead, \textsuperscript{\Letter} Corresponding author}, \textbf{Zhicheng Guo}\textsuperscript{1,2}\footnotemark[1], \textbf{Jingwen Wu}\textsuperscript{3}\footnotemark[1], \textbf{Kechen Fang}\textsuperscript{4}, \\
\textbf{Peng Li}\textsuperscript{2\ \Letter}, \textbf{Huaping Liu}\textsuperscript{1}, \textbf{Yang Liu}\textsuperscript{1,2\ \Letter} \\
\textsuperscript{1}Department of Computer Science and Technology, Tsinghua University \\
\textsuperscript{2}Institute for AI Industry Research (AIR), Tsinghua University \\
\textsuperscript{3}Department of Electrical and Computer Engineering, University of Toronto \\
\textsuperscript{4}Zhili College, Tsinghua University
\textsuperscript{5}01.AI \\
\texttt{csj23@mails.tsinghua.edu.cn} \\
}

\begin{document}
\maketitle
\input{sec/0_abstract}    
\input{sec/1_intro}

\input{sec/2_related}

\input{sec/3_VLEAI}
\input{sec/6_result}
\input{sec/7_analysis}
\input{sec/8_conclusion}
{
    \small
    \bibliographystyle{ieeenat_fullname}
    \bibliography{main}
}

\input{sec/X_suppl}

\end{document}

%% file: preamble.tex
\usepackage[dvipsnames]{xcolor}


%% file: sec/0_abstract.tex
\begin{abstract}
Vision-language models (VLMs) have recently shown promising results in traditional downstream tasks.
Evaluation studies have emerged to assess their abilities, with the majority focusing on the third-person perspective, and only a few addressing specific tasks from the first-person perspective.
However, the capability of VLMs to ``think'' from a first-person perspective, a crucial attribute for advancing autonomous agents and robotics, remains largely unexplored. 
To bridge this research gap, we introduce EgoThink, a novel visual question-answering benchmark that encompasses six core capabilities with twelve detailed dimensions.
The benchmark is constructed using selected clips from egocentric videos, with manually annotated question-answer pairs containing first-person information. 
To comprehensively assess VLMs, we evaluate twenty-one popular VLMs on EgoThink. 
Moreover, given the open-ended format of the answers, we use GPT-4 as the automatic judge to compute single-answer grading.
Experimental results indicate that although GPT-4V leads in numerous dimensions, all evaluated VLMs still possess considerable potential for improvement in first-person perspective tasks.
Meanwhile, enlarging the number of trainable parameters has the most significant impact on model performance on EgoThink.
In conclusion, EgoThink serves as a valuable addition to existing evaluation benchmarks for VLMs, providing an indispensable resource for future research in the realm of embodied artificial intelligence and robotics.
\blfootnote{Project page: \href{https://adacheng.github.io/EgoThink/}{https://adacheng.github.io/EgoThink/}}
\blfootnote{GitHub page: \href{https://github.com/AdaCheng/EgoThink}{https://github.com/AdaCheng/EgoThink/}}
\blfootnote{Dataset page: \href{https://huggingface.co/datasets/EgoThink/EgoThink}{https://huggingface.co/datasets/EgoThink/EgoThink/}}

\end{abstract}

%% file: sec/1_intro.tex
\section{Introduction}
\label{sec:intro}

Benefiting from the rapid development of large language models (LLMs)~\citep{openai2023gpt4, brown2020language, touvron2023llama}, vision-language models (VLMs)~\citep{yang2023dawn, alayrac2022flamingo, li2023blip, driess2023palm} have shown remarkable progress in both conventional vision-language downstream tasks~\citep{yang2023dawn, alayrac2022flamingo, li2023blip, driess2023palm} and following diverse human instructions~\citep{instructblip, li2023otter, ye2023mplugowl, zhu2023minigpt, liu2023llava}. Their application has expanded into broader domains such as  robotics~\citep{gao2023physically, huang2023voxposer, kuo2022f} and embodied artificial intelligence (EAI)~\citep{yang2023octopus, sumers2023distilling}. As a result, the thorough evaluation of VLMs has become increasingly important and challenging.
Observing and understanding the world from a first-person perspective is a natural approach for both humans and artificial intelligence agents. We propose that the ability to ``think'' from a first-person perspective, especially when interpreting egocentric images, is crucial for VLMs. 

\begin{figure}[t!]
    \centering
    \includegraphics[width=\linewidth]{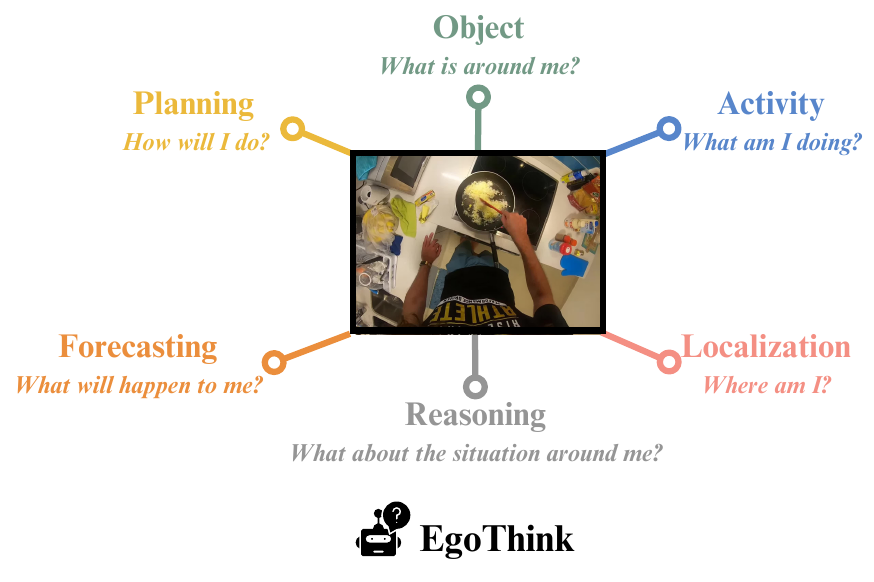}
    \caption{The main categories of our EgoThink benchmark to comprehensively assess the capability of thinking from a first-person perspective.}
    \label{fig:intro}
\end{figure}


\begin{table*}[t!]
    \centering
    \resizebox{\linewidth}{!}{
    \begin{tabular}{lcccccc}
    \toprule
        \textbf{Benchmark} & \textbf{Capability} & \textbf{Perspective} & \textbf{Data Source} & \textbf{Answer Type} & \textbf{Evaluator} & \textbf{Size} \\
    \midrule
        VL-CheckList~\citep{zhao2022vl} & Object / Attribute / Relation & Third & Datasets & PS & Accuracy & 410k  \\
        LVLM-eHub~\citep{xu2023lvlm} & General Multi-Modality & Third & Datasets & MC / OE & Metrics / LLMs / User & 332k \\
        MME~\citep{fu2023mme} & General Multi-Modality & Third & Handcraft & MC & Accuracy & 2,194 \\
        Tiny LVLM-eHub~\citep{shao2023tiny} & General Multi-Modality & Third & Datasets & OE & LLMs & 2,100 \\
        MMBench~\citep{liu2023mmbench} & General Multi-Modality & Third & Datasets / Handcraft / LLMs & MC & LLMs & 2,974 \\
        PCA-EVAL~\citep{chen2023endtoend} & Decision-Making & Third  & Handcraft & OE & Accuracy / User & 300 \\
        EgoTaskQA ~\citep{jia2022egotaskqa} & Spatial / Temporal / Causal & First & Crowdsourcing & OE & Crowdsourcing & 40k \\
        EgoVQA~\citep{fan2019egovqa} & Object / Action / Person & Third / First & Handcraft & MC & Accuracy & 520 \\
        \textbf{EgoThink} (Ours) & First-Person Thinking & First & Handcraft & OE & LLMs & 700 \\
    \bottomrule
    \end{tabular}}
    \caption{Comparison of recent comprehensive evaluation benchmarks of VLMs and our proposed benchmark EgoThink. Third and first indicate third-person and first-person perspectives. Datasets/Handcraft/LLMs denote existing datasets, manual annotation, and automatic generation by LLMs. PS/MC/OE indicate pairwise scoring, multi-choice, and open-ended question-answering, respectively.}
    \label{tab:comparison}
\end{table*}

However, as shown in Table~\ref{tab:comparison}, the ability to think from a first-person perspective is not adequately addressed by current evaluation benchmarks for VLMs. On one hand, most of these benchmarks (six out of nine, as listed in Table~\ref{tab:comparison}) focus solely on the third-person perspective. On the other hand, those benchmarks that do consider the first-person perspective only encompass a limited range of capabilities. For instance, EgoTaskQA~\citep{jia2022egotaskqa} examines spatial, temporal, and causal aspects, whereas EgoVQA~\citep{fan2019egovqa} is limited to object, action, and person aspects. Therefore, there is a clear need to develop a comprehensive benchmark to evaluate the first-person capabilities of VLMs more effectively.

In this work, we introduce a new benchmark for VLMs from a first-person perspective, named EgoThink. The initial step in developing this benchmark involves determining the necessary capabilities to assess. Humans, when interacting with the real world, consider a series of questions centered on themselves, ranging from ``{\it What is around me?}'', ``{\it What am I doing?}'', ``{\it Where am I?}'', ``{\it What about the situation around me?}'', ``{\it What will happen to me?}'' to ``{\it How will I do?}''. Drawing inspiration from this, we evaluate six core capabilities of VLMs, namely object, activity, localization, reasoning, forecasting, and planning. Each capability corresponds to one of the aforementioned questions, as illustrated in Figure~\ref{fig:intro}. The next step is constructing the benchmark. We first categorize the six core capabilities into twelve detailed dimensions. We then select a minimum of 50 distinct and clear clips from egocentric videos for each dimension and manually annotate them with relevant first-person question-answer pairs. This approach ensures the quality and variety of the benchmark. The final step is evaluating VLM performance on this benchmark. Building on recent studies~\citep{vicuna2023, shao2023tiny, berrios2023towards}, we use GPT-4~\citep{openai2023gpt4} as an automatic evaluator. {The Pearson correlation coefficient, when compared with human evaluation, shows a value of 0.68, indicating that the evaluation results are dependable.}

Based on our proposed EgoThink benchmark, we conduct comprehensive experiments to evaluate the first-person capabilities of twenty-one popular VLMs with varying model and data compositions. The findings indicate that GPT-4V stands out as the most effective model in various aspects. However, it shows less impressive results in specific capabilities such as activity and counting. Additionally, we observed that no single VLM consistently surpasses others in every aspect. For instance, GPT-4V is less effective than BLIP-2-11B for localization. Increasing the language model portion of the VLMs generally leads to better performance, but this improvement is not uniform across all models. Finally, our results highlight a significant potential for further enhancing the first-person capabilities of VLMs.


%% file: sec/2_related.tex
\section{Related Work}
\label{sec:related}

\begin{figure*}[t!]
    \centering
    \includegraphics[width=\linewidth]{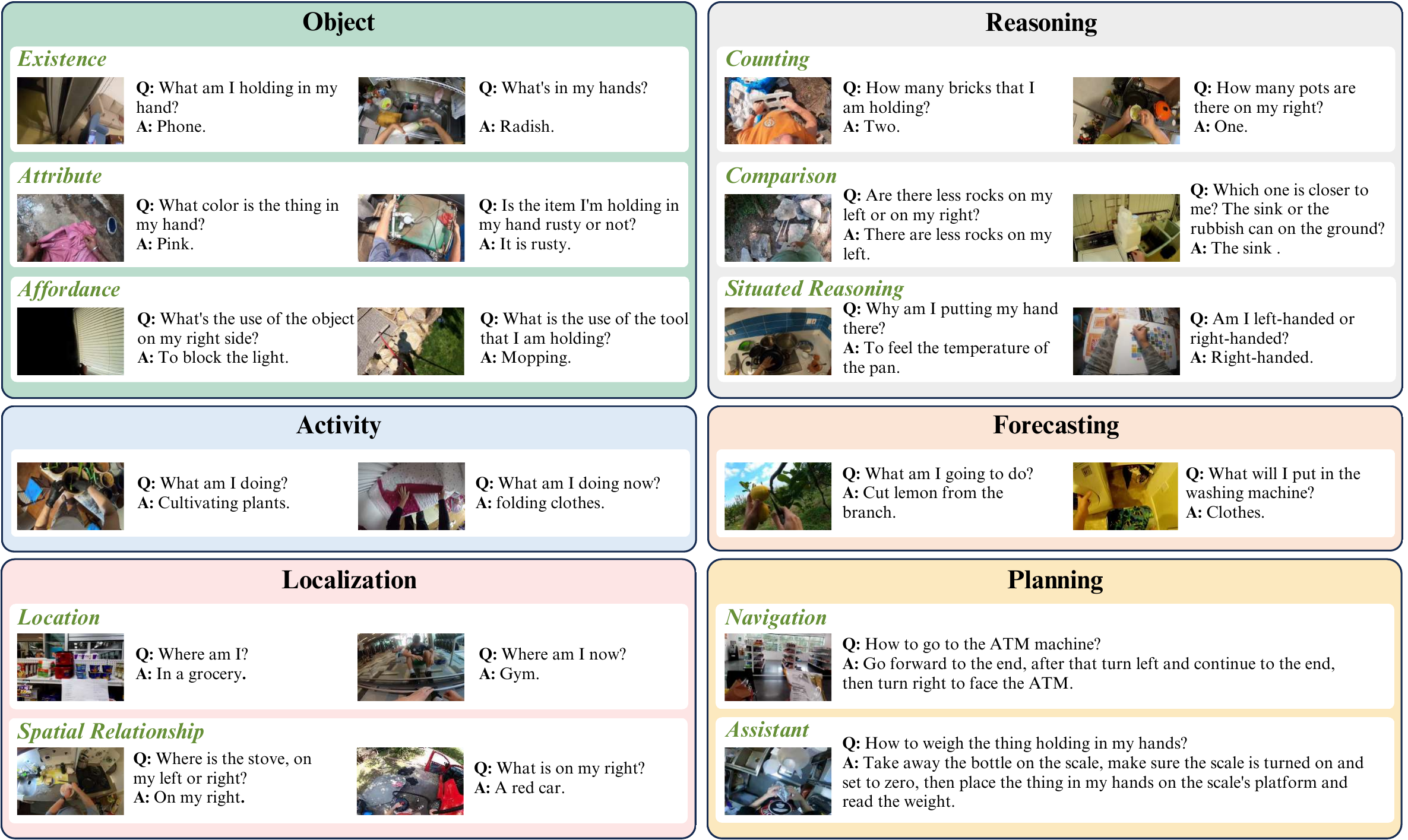}
    \caption{Categories with fine-grained dimensions and their corresponding examples of EgoThink benchmark.}
    \label{fig:overall}
\end{figure*}

\noindent\textbf{Vision-Language Models.}
Inspired by the impressive success of LLMs~\citep{brown2020language, ouyang2022training, wang2023openchat}, the recent popular VLMs tend to regard the powerful LLMs as the core backbone.
At the beginning, VLMs usually use large-scale image-text pairwise datasets~\citep{jia2021scaling, changpinyo2021conceptual, lin2014microsoft} or arbitrarily interleaved visual and textual data~\citep{zhu2023multimodal, alayrac2022flamingo} to pre-train.
Furthermore, thanks to the availability of enormous image-text instruction datasets~\citep{liu2023visual, li2023mimic}, recent studies~\citep{instructblip, li2023otter, ye2023mplugowl, zhu2023minigpt, liu2023llava} further apply instruction tuning to help VLMs generate satisfactory answers.
Benefiting from the two-stage training process, recent VLMs can achieve stunning performance on downstream vision-language tasks~\citep{lin2014microsoft, plummer2015flickr30k, antol2015vqa, hudson2019gqa}.

\noindent\textbf{Evaluations of VLMs.}
To evaluate the abilities of VLMs, there are diverse types of vision language downstream tasks.
Conventional benchmarks, such as image caption tasks~\citep{hossain2019comprehensive, you2016image} and visual question reasoning tasks~\citep{goyal2017making, schwenk2022okvqa}, mainly probe specific abilities of VLMs from the third-person perspective.
Meanwhile, specialized analytical studies comprehensively evaluate the performance of VLMs from the third-person perspective, where Vlue~\citep{zhou2022vlue} consists of five fundamental tasks and Lvlm-ehub~\citep{xu2023lvlm} evaluates six categories of capabilities on 47 standard vision-language benchmarks. 
As for the first-person perspective, there are some egocentric evaluation benchmarks in the computer vision field to assess some visual capabilities~\citep{liu2022hoi4d, zhu2023egoobjects, zhang2018egogesture, huang2016pointing}.
In terms of multi-modality, there are a few benchmarks, such as EgoVQA\citep{fan2019egovqa} and EgoTaskQA~\citep{jia2022egotaskqa}, where mainly specific tasks without an overall understanding.
In this paper, we mainly focus on exploring the comprehensive capabilities of VLMs to think from a first-person perspective, as a supplement to previous evaluation benchmarks.

%% file: sec/3_VLEAI.tex
\section{EgoThink Benchmark}

In this section, we first elaborate on the core capabilities of thinking from a first-person perspective. 
Then, we introduce the process to manually construct our proposed benchmark EgoThink, which asks VLMs to generate open-ended answers according to first-person images and questions.


\subsection{Core Capabilities}


As shown in Figure~\ref{fig:overall}, we specifically design six categories with twelve fine-grained dimensions from the first-person perspective for quantitative evaluation.

\begin{itemize}
    \item \textbf{Object: What is around me?} Recognizing objects in the real world is a preliminary ability of the human visual system~\citep{zou2023object, liu2020deep, zhao2019object}. Images from a first-person or egocentric perspective~\citep{zhu2023egoobjects, ramanathan2023paco, liu2022hoi4d} pay more attention to the objects surrounding the subject or in hands. Moreover, we further divide the object category into three fine-grained dimensions: 
    (1) \textit{Existence}, predicting whether there is an object as described in the images;
    (2) \textit{Attribute}~\citep{farhadi2009describing, kanwisher1992objects}, detecting properties or characteristics (e.g., color) of an object; 
    (3) \textit{Affordance}~\citep{hermans2011affordance, montesano2008learning}, predicting potential actions that a human can apply to an object.

    \item \textbf{Activity: What am I doing?} Activity recognition is to automatically recognize specific human activities in video frames or still images~\citep{vrigkas2015review, ke2013review, jobanputra2019human}. From the egocentric perspective, we mainly focus on actions or activities based on object-hand interaction~\citep{fathi2011understanding, bambach2015lending, nguyen2016recognition}.
    \item \textbf{Localization: Where am I?} In reality, localization is a critical capability for navigation and scene understanding in the real world~\citep{montemerlo2007fastslam, saeedi2016multiple}. Here we investigate the localization capability from two aspects, \textit{Location} and \textit{Spatial Relationship}. Location indicates detecting the scene surrounding the subject~\citep{do2022learning, hayet2003visual}. Spatial reasoning contains allocentric and egocentric perspectives~\citep{klatzky1998allocentric, mou2004allocentric, morganti2013allo, grade2015evidence}. We focus on the egocentric perspective, i.e., the position of the object with respect to the subject.
\end{itemize}

\begin{table*}[t!]
    \centering
    \resizebox{\textwidth}{!}{
    \begin{tabular}{lccccccccc}
    \toprule
        \multirow{2}{*}{\textbf{VLMs}} &
        \multirow{2}{*}{\textbf{Image Encoder}} & \multirow{2}{*}{\textbf{LLM}} & \multirow{2}{*}{\textbf{Alignment Module}} & \multirow{2}{*}{\textbf{TTP}} & \multirow{2}{*}{\textbf{ToP}} &
        \multicolumn{2}{c}{\textbf{Dataset Size}} & \multirow{2}{*}{\textbf{EgoData}} & \multirow{2}{*}{\textbf{Video}} \\
        & & & & & & Image-Text & Instruction\\
        \midrule
        \rowcolor[gray]{0.95} \multicolumn{10}{c}{\textbf{API-based Model}} \\ 
        \midrule
        \textbf{GPT-4V~\citep{openai2023gpt4}} & \multicolumn{8}{c}{Unknown} \\
        \midrule
        \rowcolor[gray]{0.95} \multicolumn{10}{c}{\textbf{$\sim$7B Models}} \\
        \midrule
        \textbf{OpenFlamingo-7B~\citep{alayrac2022flamingo, awadalla2023openflamingo}} & CLIP-ViT-L & $\text{MPT}_{\text{7B}}$ & Attention & 1.4B & 8.1B & 2B & - & \XSolidBrush & \Checkmark \\
        \textbf{BLIP-2-6.7B~\citep{li2023blip}} & EVA-CLIP-ViT-g &  $\text{OPT}_{\text{6.7B}}$ & Q-Former & 108M & 7.8B & 129M & - & \XSolidBrush & \XSolidBrush \\
        \textbf{VideoChat-7B~\citep{li2023videochat}} & BLIP2-VE & $\text{Vicuna}_\text{7B}$ & Q-Former & 205M & 8B & 25M & 18K & \XSolidBrush & \Checkmark \\
        \textbf{LLaVA-1.5-7B~\citep{liu2023improved}} & CLIP-ViT-L-336px & $\text{Llama2}_\text{7B}$ & MLP & 6.8B & 7.1B  & 558k & 665k & \XSolidBrush & \XSolidBrush \\
        \textbf{MiniGPT-4-7B~\citep{zhu2023minigpt}} & BLIP2-VE & $\text{Llama2}_{\text{7B}}$ & Linear & 23M & 7.7B & 5M & 3.5k & \XSolidBrush & \XSolidBrush\\
        \textbf{InstructBLIP-7B~\citep{instructblip}} & EVA-CLIP-ViT-g & $\text{Vicuna}_{\text{7B}}$ & Q-Former & 189M & 7.9B & - & 16M & \XSolidBrush & \XSolidBrush \\
        \textbf{LLaMA-Adapter-7B~\citep{gao2023llama}} & CLIP-ViT-L & $\text{LLaMA}_\text{7B}$ & Early Fusion & 14M & 7.2B & 567k & 52k & \XSolidBrush & \XSolidBrush \\
        \textbf{Otter-I-7B~\citep{li2023otter}} & CLIP-ViT-L & $\text{MPT}_\text{7B}$ & Attention & 1.4B & 8.1B & - & 2.8B & \XSolidBrush & \Checkmark \\
        \textbf{PandaGPT-7B~\citep{su2023pandagpt}} & ImageBind & $\text{Vicuna}_\text{7B}$ & Linear + LLM LoRA & 38M & 7.9B & - & 160k & \Checkmark & \Checkmark \\
        \textbf{mPLUG-owl-7B~\citep{ye2023mplugowl}} & CLIP-ViT-L & $\text{LLaMA}_\text{7B}$ & Attention & 4M & 7.1B & 204M & 158k & \XSolidBrush & \XSolidBrush \\
        \textbf{Video-LLaVA-7B~\citep{lin2023video}} & LanguageBind & $\text{Vicuna}_\text{7B}$ & Linear  & 6.8B & 7.5B & 1260k & 765k & \XSolidBrush & \Checkmark \\
        \textbf{LLaVA-7B~\citep{liu2023llava}} & CLIP-ViT-L & $\text{Llama2}_\text{7B}$ & Linear & 6.7B & 7.1B & 595k & 158k & \XSolidBrush & \XSolidBrush \\
        \textbf{ShareGPT4V-7B}~\citep{chen2023sharegpt4v} & CLIP-ViT-L-336px & $\text{Vicuna}_\text{7B}$ & MLP & 6.7B & 6.7B & 1.2M & 665k & \XSolidBrush & \XSolidBrush  \\
        \midrule
        \rowcolor[gray]{0.95} \multicolumn{10}{c}{\textbf{$\sim$13B Models}} \\
        \midrule
        \textbf{InstructBLIP-13B~\citep{instructblip}} & EVA-CLIP-ViT-g & $\text{Vicuna}_\text{13B}$ & Q-Former & 189M & 14.2B &  - & 16M & \XSolidBrush & \XSolidBrush \\
        \textbf{PandaGPT-13B~\citep{su2023pandagpt}} & ImageBind & $\text{Vicuna}_\text{13B}$ & Linear+LLM LoRA & 52M & 13.1B & - & 160k & \Checkmark & \Checkmark \\
        \textbf{LLaVA-13B-Vicuna~\citep{liu2023llava}} & CLIP-ViT-L-336px & $\text{Vicuna}_\text{13B}$ & Linear & 13.0B & 13.3B &  595k & 158k & \XSolidBrush & \XSolidBrush \\
        \textbf{BLIP-2-11B~\citep{li2023blip}} & EVA-CLIP-ViT-g & $\text{FlanT5}_{\text{XXL}}$ & Q-Former & 108M & 12.2B & 129M & - & \XSolidBrush & \XSolidBrush\\
        \textbf{InstructBLIP-11B~\citep{instructblip}} & EVA-CLIP-ViT-g & $\text{FlanT5}_{\text{XXL}}$ & Q-Former & 189M & 12.3B &  - & 16M & \XSolidBrush & \XSolidBrush \\
        \textbf{LLaVA-13B-Llama2~\citep{liu2023llava}} & CLIP-ViT-L & $\text{Llama2}_\text{13B}$ & Linear & 13.0B & 13.3B &  595k & 158k & \XSolidBrush & \XSolidBrush\\
        \textbf{LLaVA-1.5-13B~\citep{liu2023improved}} & CLIP-ViT-L-336px & $\text{Llama2}_\text{13B}$ & MLP & 13.0B & 13.4B &  558k & 665k & \XSolidBrush & \XSolidBrush \\


    \bottomrule
    \end{tabular}}
    \caption{Statistics of compared API-based and open-source VLMs, where TTP and ToP indicate Total Trainable Parameters and Total Parameters, respectively. Moreover, EgoData and Video indicate that there are egocentric visual data and video data for training, respectively.} 
    \label{tab:model_sta}
\end{table*}

\begin{itemize}
    \item \textbf{Reasoning: What about the situation around me?} During the complex decision-making process, reasoning lies everywhere in our lives. 
    Here we mainly focus on \textit{Counting}, \textit{Comparison}, and \textit{Situated Reasoning}.
    Due to the first-person perspective, we generally count or compare objects in our hands or surrounding ourselves. As for situated reasoning, we employ cases that cannot be answered directly from the information in the images and require further reasoning processes.
    \item \textbf{Forecasting: What will happen to me?} Forecasting~\citep{liu2022joint, liu2020forecasting, furnari2017next, grauman2022ego4d} is a critical skill in the real world. From an egocentric view, forecasting always predicts the future of object-state transformation or hand-object interactions.
    \item \textbf{Planning: How will I do?} In reality, planning~\citep{huang2022inner, ahn2022can, song2023llm} is an important capability to deal with complex problems, typically applied in \textit{Navigation}~\citep{pandey2014path, patle2019review, susperregi2012interacting} and \textit{Assistance}~\citep{wojtak2021neural, helms2002rob}. Navigation is going to a goal location from a start position, while assistance is offering instructions to solve daily problems.
\end{itemize}

\subsection{Data Collection}

In this section, we mainly introduce the detailed processing to construct our EgoThink benchmark.

\noindent\textbf{Collecting first-person visual data.}
Firstly, we leverage a popular and large egocentric video dataset Ego4D~\citep{grauman2022ego4d}, which is designed to advance the field of first-person perception in computer vision. 
To obtain a diverse representation in different scenarios, Ego4D encompasses 3,670 hours of video from 931 unique camera wearers spanning 74 global locations across 9 countries.
To collect first-person visual data, we begin by extracting every frame from a subset of the Ego4D video dataset, yielding a diverse raw image dataset.
Please note that our current focus is solely on images, as most VLMs today do not support video input. We intend to expand our scope to include videos in our future work.
Considering the heavy human labor and the diversity of scenarios, we sample images every few dozen frames.
To ensure high quality, we apply strict criteria for selecting the extracted frames.
We first exclude images that lack clarity or fail to exhibit egocentric characteristics.
Then, to obtain the high diversity within the dataset, we conduct a further screening to ensure that at most two images per video are included in the filtered image set.
Finally, we obtain enormous high-quality images with exhibit egocentric characteristics as first-person image candidates.
\begin{figure}[t!]
    \centering
    \includegraphics[width=0.9\linewidth]{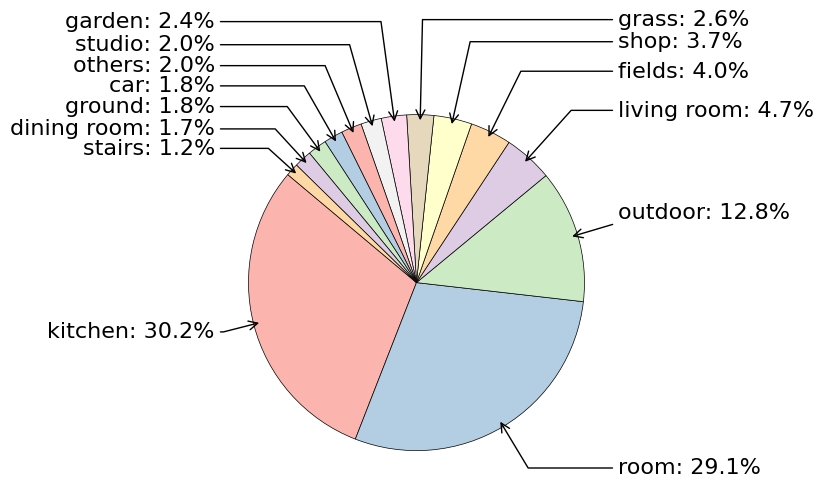}
    \caption{This chart illustrates the distribution of various scene categories within the EgoThink dataset. The `others' category encompasses 13 different scene types, each representing less than one percent of total scenes.}
    \label{fig:scenes}
\end{figure}

\begin{table*}[t!]
    \centering
    \resizebox{\textwidth}{!}{
    \begin{tabular}{l|ccc|c|cc|ccc|c|cc|c}
    \toprule
        \multirow{2}{*}{\textbf{Methods}} & \multicolumn{3}{c|}{\textbf{Object}} & \multirow{2}{*}{\textbf{Activity}} & \multicolumn{2}{c|}{\textbf{Localization}} & \multicolumn{3}{c|}{\textbf{Reasoning}} & \multirow{2}{*}{\textbf{Forecasting}} & \multicolumn{2}{c|}{\textbf{Planning}} & \multirow{2}{*}{\textbf{Average}} \\ [0.2em]
         & {Exist} & {Attr} & {Afford} &  & {Loc} & {Spatial} & {Count} & {Compar} & {Situated} & & {Nav} & {Assist} \\ 
        \toprule 
        \rowcolor[gray]{0.95} \multicolumn{14}{c}{\textbf{API-based model}} \\ 
        \midrule
        \textbf{GPT-4V} & 62.0 & \textbf{82.0} & \textbf{58.0} & \textbf{59.5} & \underline{86.0} & \underline{62.0} & \textbf{42.0} & 48.0 & \textbf{83.0} & \textbf{55.0} & \textbf{64.0} & \textbf{84.0} & \textbf{65.5} \\
        \midrule 
        \rowcolor[gray]{0.95} \multicolumn{14}{c}{\textbf{$\sim$7B Models}} \\
        \midrule
        \textbf{OpenFlamingo-7B} & 16.0 & 55.0 & 37.0 & 15.0 & 34.0 & 34.0 & 21.0 & 40.0 & 21.0 & 31.0 & 11.0 & 11.0 & 27.2 \\
        \textbf{BLIP-2-6.7B} & 49.0 & 29.0 & 39.0 & 33.5 & 60.0 & 31.0 & \ \ 3.0 & 21.0 & 33.0 & 25.0 & \ \ 8.0 & \ \ 6.0 & 28.1 \\
        \textbf{VideoChat-7B} & 46.0 & 44.0 & 36.0 & 45.0 & 61.0 & 42.0 & 36.0 & 39.0 & 32.0 & 26.5 & 13.0 & 21.0 & 36.8 \\
        \textbf{LLaVA-1.5-7B} & 33.0 & 47.0 & \underline{54.0} & 35.5 & 35.0 & 49.0 & 20.0 & 47.0 & 37.0 & 27.0  & 29.0 & 54.0 & 39.0 \\
        \textbf{MiniGPT-4-7B} & 50.0 & 56.0 & 46.0 & 39.0 & 55.0 & 49.0 & 14.0 & 48.0 & 31.0 & 41.5 & 14.0 & 44.0 & 40.6 \\
        \textbf{InstructBLIP-7B} & 50.0 & 33.0 & 45.0 & 47.5 & 77.0 & 38.0 & 18.0 & 43.0 & 67.0 & 40.5 & 19.0 & 31.0 & 42.4 \\
        \textbf{LLaMA-Adapter-7B} & 37.0 & 60.0 & 46.0 & 34.5 & 48.0 & 51.0 & 29.0 & 39.0 & 25.0 & 41.5 & 42.0 & 57.0 & 42.5 \\
        \textbf{Otter-I-7B} & 48.0 & 56.0 & 39.0 & 44.0 & 60.0 & 44.0 & \underline{39.0} & 48.0 & 42.0 & 38.0 & 31.0 & 55.0 & 45.3 \\
        \textbf{PandaGPT-7B} & 40.0 & 56.0 & 41.0 & 37.0 & 61.0 & 52.0 & 19.0 & \underline{52.0} & 53.0 & 43.0 & 39.0 & 61.0 & 46.2 \\
        \textbf{mPLUG-owl-7B} & 56.0 & 58.0 & 47.0 & 53.0 & 60.0 & 53.0 & 25.0 & 49.0 & 44.0 & 49.5 & 33.0 & 58.0 & 48.8  \\
        \textbf{Video-LLaVA-7B} & 56.0 & 60.0 & 53.0 & 45.0 & \underline{86.0} & 60.0 & \underline{39.0} & 38.0 & 60.0 & 46.5 & 11.0 & 38.0 & 49.4 \\
        \textbf{LLaVA-7B} & 63.0 & 58.0 & 50.0 & 47.0 & 81.0 & 45.0 & 24.0 & 36.0 & 47.0 & 49.5 & 35.0 & 60.0 & 49.6 \\
        \textbf{ShareGPT4V-7B} & \underline{67.0} & \underline{75.0} & 53.0 & 55.5 & 77.0 & \underline{62.0} & 30.0 & 38.0 & 66.0 & 47.0 & 41.0 & 63.0 & 51.9 \\
        \midrule 
        \rowcolor[gray]{0.95} \multicolumn{14}{c}{\textbf{$\sim$13B Models}} \\
        \midrule
        \textbf{InstructBLIP-13B} & 52.0 & 55.0 & 49.0 & 54.0 & 63.0 & 49.0 & 11.0 & 33.0 & 59.0 & 44.0 & 19.0 & 25.0 & 42.8 \\
        \textbf{PandaGPT-13B} & 35.0 & 52.0 & 41.0 & 40.5 & 68.0 & 31.0 & 32.0 & 40.0 & 47.0 & 45.5 & 16.0 & {69.0} & 43.1 \\
        \textbf{LLaVA-13B-Vicuna} & 54.0 & 62.0 & 52.0 & 46.0 & 53.0 & 46.0 & 26.0 & 44.0 & 29.0 & 44.0 & 35.0 & 66.0 & 46.4 \\
        \textbf{BLIP-2-11B} & 52.0 & 62.0 & 41.0 & 49.5 & \textbf{90.0} & \textbf{66.0} & 25.0 & 50.0 & 70.0 & 48.0 & 18.0 & 24.0 & 49.6 \\
        \textbf{InstructBLIP-11B} & \textbf{74.0} & {68.0} & 48.0 & 49.5 & \underline{86.0} & 52.0 & 32.0 & {49.0} & \underline{73.0} & \underline{53.0} & 16.0 & 17.0 & 51.5 \\
        \textbf{LLaVA-13B-Llama2} & 65.0 & 61.0 & 45.0 & \underline{56.0} & 77.0 & 53.0 & 34.0 & 34.0 & 66.0 & 50.5 & \underline{49.0} & \underline{71.0} & 55.1\\
        \textbf{LLaVA-1.5-13B} & {66.0} & 55.0 & 51.0 & 55.0 & 82.0 & 57.0 & 32.0 & \textbf{56.0} & 67.0 & 48.5 & 39.0 & 55.0 & \underline{55.3} \\         
    \bottomrule
    \end{tabular}}
    \caption{Combined single-answer grading scores on zero-shot setups for various dimensions. The \textbf{bold} indicates the best performance while the \underline{underline} indicates the second-best performance. Exist, Attr, Afford, Loc, Spatial, Count, Compar, Situated, Nav and Assist represent existence, attribute, affordance, location, spatial relationship, counting, comparison, situated reasoning, navigation, and assistance.} 
    \label{tab:combined_grading}
\end{table*}


\noindent\textbf{Annotating question-answer pairs.}
Upon receiving a substantial collection of first-person image candidates, we engage six annotators to manually label question-answer pairs.
Given that the EgoThink benchmark is composed of twelve dimensions, annotators were responsible for two specific dimensions.
The annotators can access all the image candidates and are asked to select appropriate images to annotate their corresponding question-answer pairs to relevant categories.
Once the image is selected, it will be removed from the candidates to ensure no repetition.
Moreover, to ensure the correctness of our annotations, we have three additional annotators to review the question-answer pairs after the first annotation process.
The annotation will not be reserved until the three annotators all agree that the first-person visual data and the assigned question-answer pairs meet the definition of a specific dimension.

\noindent\textbf{Statistics.} The EgoThink benchmark comprises a collection of 700 images across six categories with twelve fine-grained dimensions. These images are extracted from 595 videos, ensuring a broad representation of scenarios. To guarantee diversity, a wide range of scenes and concepts has been deliberately selected. As depicted in Figure~\ref{fig:scenes}, the dataset encompasses a diverse range of scenes, covering key scenarios relevant to EAI. Furthermore, we have meticulously crafted question and answer for each image in the EgoThink benchmark, aiming to closely replicate real-life conversations. This involves employing different question types, varying questions in length and complexity, paired with well-reasoned and accurate answers. 
In addition, the size of the dataset is a well-balanced trade-off between benchmark diversity and high cost of open-ended QA evaluation. We ensure our 700 examples were diverse in concepts, scenes, and videos to provide a robust performance estimation within practical limits.
Detailed statistics of the EgoThink benchmark are presented in Appendix~\ref{app:statistics}.

%% file: sec/6_result.tex
\section{Experiments}

\begin{figure}[t!]
\centering
\footnotesize
\begin{tcolorbox}[
  enhanced,        
  colback=white,   
  colframe=black,  
  arc=3mm,         
  width=\columnwidth, 
  boxrule=0.8pt,    
  left=5pt,  
  right=1pt  
]
\begin{tabular}{m{0.30\columnwidth}p{0.59\columnwidth}}
\toprule
\cellcolor{myblue!40}\textbf{Image} & \cellcolor{myblue!40}\textbf{Original Question \& Answer} \\ 
\midrule
\addlinespace[1mm]
\includegraphics[width=0.30\columnwidth]{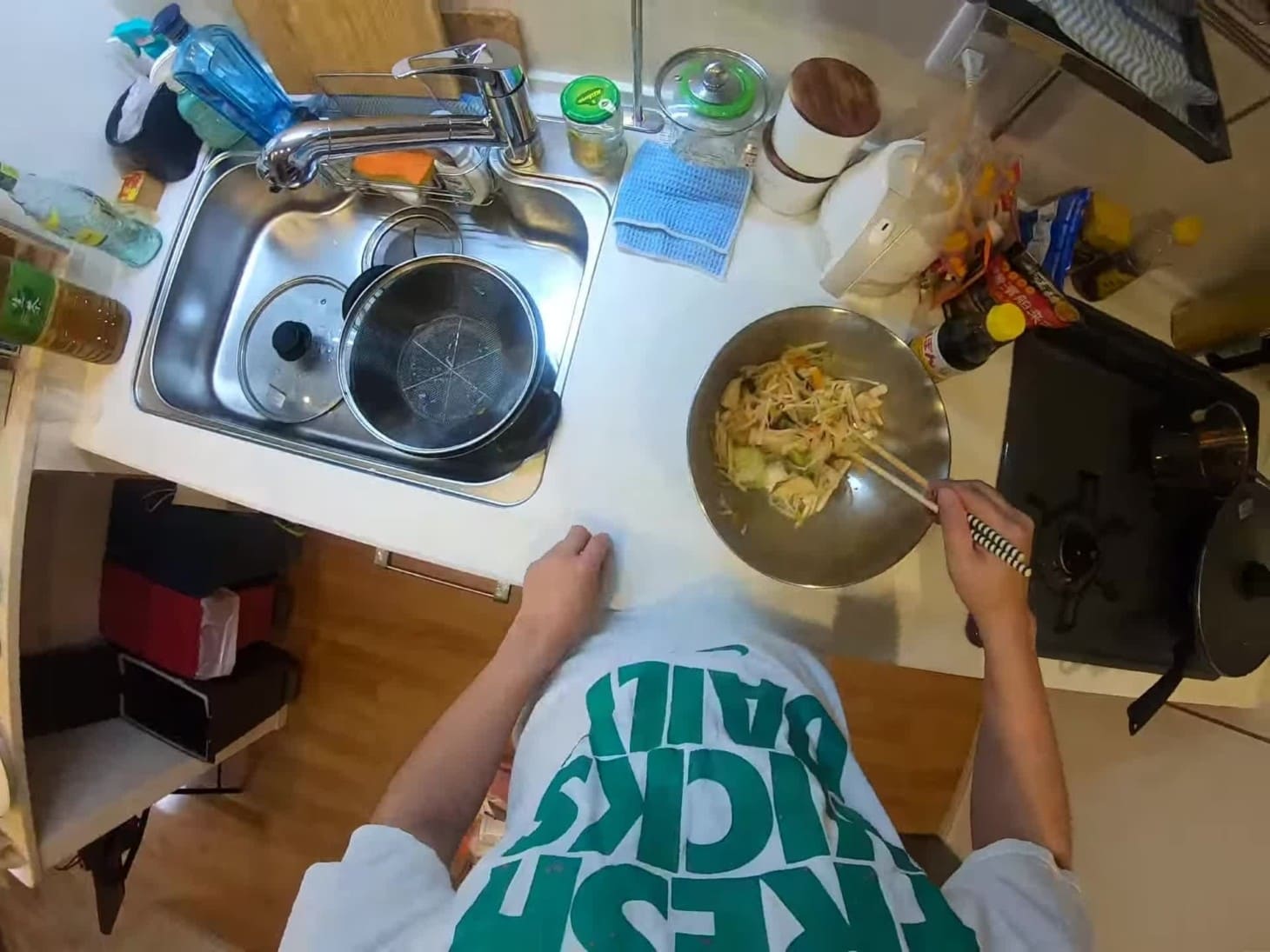} & \makecell[l]{\textit{Question:} Am I holding chopsticks in my \\ right hand?
 \\ \\ \textit{Answer:} Yes.} \\
\midrule
\multicolumn{2}{l}{\cellcolor{red!20}\textbf{Model Predictions}}\\
\midrule
\addlinespace[1mm]
\multicolumn{2}{m{0.95\columnwidth}}{\textbf{\texttt{GPT-4V:}} No, (you are not holding chopsticks in your right hand. It looks like you are holding \textcolor{red!80}{a pair of tongs.})} \\
\multicolumn{2}{m{0.95\columnwidth}}{\textbf{\texttt{InstructBLIP-11B:}} Yes.} \\
\multicolumn{2}{m{0.95\columnwidth}}{\textbf{\texttt{LLaVA-7B:}} Yes, you are holding chopsticks in your right hand.} \\
\midrule
\multicolumn{2}{l}{\cellcolor{myblue!40}\textbf{Changed Question Format \& Answer}}\\
\midrule
\addlinespace[1mm]
\multicolumn{2}{m{0.95\columnwidth}}{\textit{Question:} What am I holding in my right hand?} \\ 
\multicolumn{2}{m{0.95\columnwidth}}{Answer: Chopsticks.} \\
\midrule
\multicolumn{2}{l}{\cellcolor{red!20}\textbf{Model Predictions}}\\
\midrule
\addlinespace[1mm]
\multicolumn{2}{m{0.95\columnwidth}}{\textbf{\texttt{GPT-4V:}} In your right hand, you're holding what appears to be \textcolor{red!80}{a pan with a handle}, and it looks like you're cooking some food, possibly a stir-fry or sauteed vegetables.} \\
\bottomrule
\end{tabular}
\end{tcolorbox}
\caption{Case study (\textcolor{red!80}{wrong}) of GPT-4V in the existence dimension of the object ability. In both the question formats of yes/no and what, GPT-4V can not correctly detect the chopsticks in my right hand, while InstructBLIP-11B and LLaVA-7B can.}
\label{tab:existence_gpt}
\end{figure}

\subsection{Experimental Setups}

\noindent\textbf{Vision-Language Models.}
We collect the most popular eighteen types of representative VLMs to assess as shown in Table~\ref{tab:model_sta}.
Due to the possible effects of model parameters, we divide models into $\sim$7B and $\sim$13B for a fair comparison. 
Detailed information about VLMs can be found in Appendix~\ref{sec:vlms}.
We conduct zero-shot setups for all VLMs across our EgoThink benchmark. The prompts used for each VLM are shown in Appendix~\ref{app:model_inference_prompts}.

\noindent\textbf{Single-answer grading.} 
Considering evaluating open-ended model generations is not a trivial problem~\citep{shao2023tiny, liu2023mmbench}, we propose to use GPT-4~\citep{openai2023gpt4} as the automatic evaluator~\citep{zheng2023judging} to better measure generated answers. 
In this protocol, we want to measure how close one model output is to the reference. Different from traditional similarity-based methods, GPT-4 pays more attention to semantics. 
In the detailed implementation, we format the question, the model output, and the reference in a prompt as shown in Appendix~\ref{app:evaluate_prompts} and feed it into the GPT-4 evaluator. The GPT-4 evaluator is asked to assign a score of 0 (wrong), 0.5 (partially correct), or 1 (correct) to the model output.
Additionally, we further discuss to use of GPT-3.5-Turbo, Claude-2, and humans as evaluators in Section~\ref{sec:agreement}.

\subsection{Results}

We first present the overall results of the evaluated models on our EgoThink benchmark as shown in Table~\ref{tab:combined_grading}.
Despite having improved over the years, VLMs are still difficult to think from a first-person perspective, even GPT-4V.
Among the six categories, only the scores on planning and localization are relatively high, the performance in other capabilities can only reach around 60 points at best.
Among the better models, GPT-4V generally performs much better than other models, only falling short in localization dimension compared to BLIP-2-11B.
We will further introduce the detailed scores across different dimensions.
More case studies can be found in Appendix~\ref{sec:cases}.



\begin{figure}[t!]
\centering
\footnotesize
\begin{tcolorbox}[
  enhanced,        
  colback=white,   
  colframe=black,  
  arc=3mm,         
  width=\columnwidth, 
  boxrule=0.8pt,    
  left=5pt,  
  right=1pt  
]
\begin{tabular}{m{0.30\columnwidth}p{0.59\columnwidth}}
\toprule
\cellcolor{myblue!40}\textbf{Image} & \cellcolor{myblue!40}\textbf{Question \& Answer} \\ 
\midrule
\addlinespace[1mm]
\includegraphics[width=0.30\columnwidth]{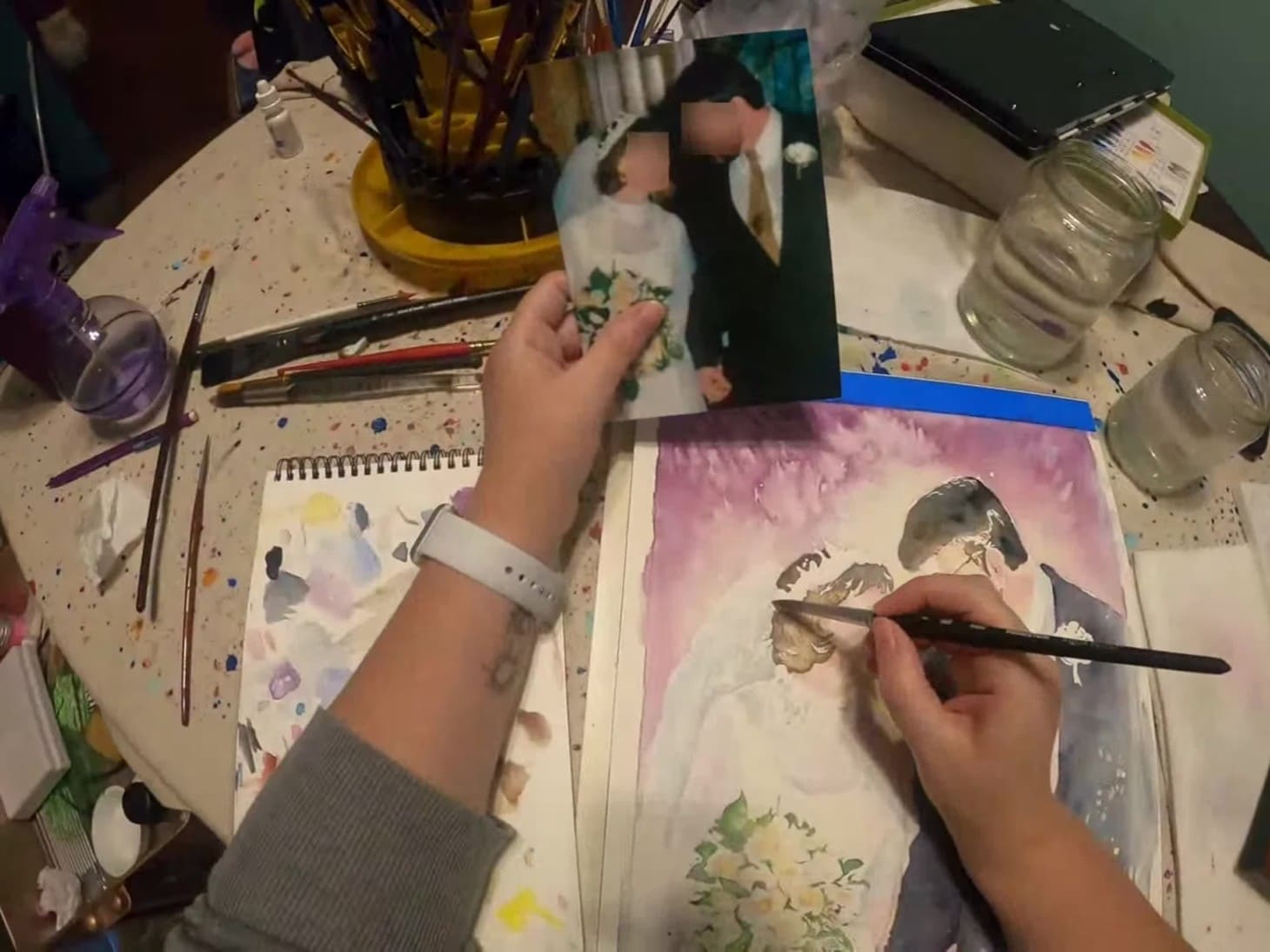} & \makecell[l]{\textit{Question:} What is the shape of the object \\ I am holding in my left hand?
\\ \\ \textit{Answer:} Rectangular.} \\
\midrule
\multicolumn{2}{l}{\cellcolor{red!20}\textbf{Model Predictions}}\\
\midrule
\addlinespace[1mm]
\multicolumn{2}{m{0.95\columnwidth}}{\textbf{\texttt{GPT-4V:}} Rectangular.} \\
\multicolumn{2}{m{0.95\columnwidth}}{\textbf{\texttt{InstructBLIP-11B:}} \textcolor{red!80}{Pen.}} \\
\multicolumn{2}{m{0.95\columnwidth}}{\textbf{\texttt{BLIP-2-11B:}} \textcolor{red!80}{A paint brush.}} \\
\multicolumn{2}{m{0.95\columnwidth}}{\textbf{\texttt{PandaGPT-13B:}} I am holding \textcolor{red!80}{a paintbrush} in my left hand.} \\

\midrule
\cellcolor{myblue!40}\textbf{Image} & \cellcolor{myblue!40}\textbf{Question \& Answer} \\ 
\midrule
\addlinespace[1mm]
\includegraphics[width=0.30\columnwidth]{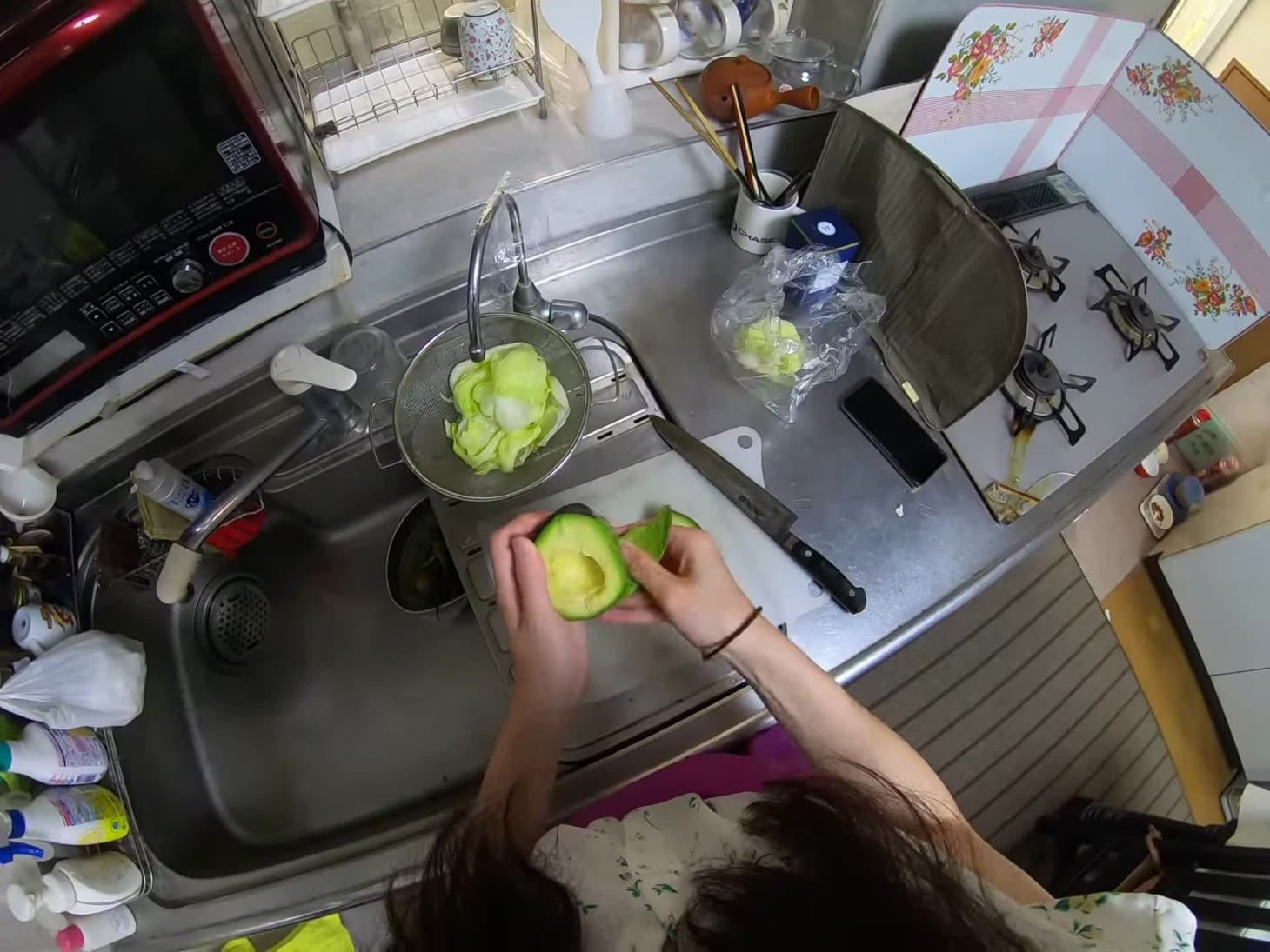} & \makecell[l]{\textit{Question:}What am I doing? \\ \\ \textit{Answer:} Peeling the avocado.} \\
\midrule
\multicolumn{2}{l}{\cellcolor{red!20}\textbf{Model Predictions}}\\
\midrule
\addlinespace[1mm]
\multicolumn{2}{m{0.95\columnwidth}}{\textbf{\texttt{GPT-4V:}}Peeling vegetables.}\\
\multicolumn{2}{m{0.95\columnwidth}}{\textbf{\texttt{BLIP-2-11B:}} \textcolor{red!80}{Slicing} an avocado.} \\
\multicolumn{2}{m{0.95\columnwidth}}{\textbf{\texttt{InstructBLIP-11B:}} \textcolor{red!80}{Chopping.}} \\
\multicolumn{2}{m{0.95\columnwidth}}{\textbf{\texttt{LLaVA-13B-Vicuna:}} You are \textcolor{red!80} {cutting up} an avocado in a kitchen sink.} \\
\bottomrule
\end{tabular}
\end{tcolorbox}
\caption{Case studies (\textcolor{red!80}{wrong}) in the attribute, affordance, and activity dimensions. The top case demonstrates some VLMs locate the wrong place in the attribute and affordance dimension. The bottom case illustrates that in the activity dimension, some VLMs fail to detect the specific action.}
\label{tab:attribute_activity}
\end{figure}

\noindent\textbf{Results on object.}
In detail, we evaluate through three dimensions, including existence, attribute, and affordance.
For existence, InstructBLIP-11B and ShareGPT4V-7B achieve the top-2 performance, indicating that they can predict the object precisely from the first-person perspective.
As for GPT-4V, as illustrated in Figure~\ref{tab:existence_gpt}, we observe that its performance in handed object detection leaves room for improvement.
As for both attribute and affordance, the GPT-4V model has demonstrated superior performance, especially in the attribute dimension.
In both dimensions, some open-source models as shown at the top of Figure~\ref{tab:attribute_activity} locate the wrong place or only answer the type of the object rather than its attribute or affordance.

\begin{figure}[t!]
\centering
\footnotesize
\begin{tcolorbox}[
  enhanced,        
  colback=white,   
  colframe=black,  
  arc=3mm,         
  width=\columnwidth, 
  boxrule=0.8pt,    
  left=5pt,  
  right=1pt  
]
\begin{tabular}{m{0.30\columnwidth}p{0.59\columnwidth}}
\toprule
\cellcolor{myblue!40}\textbf{Image} & \cellcolor{myblue!40}\textbf{Question \& Answer} \\ 
\midrule
\addlinespace[1mm]
\includegraphics[width=0.30\columnwidth]{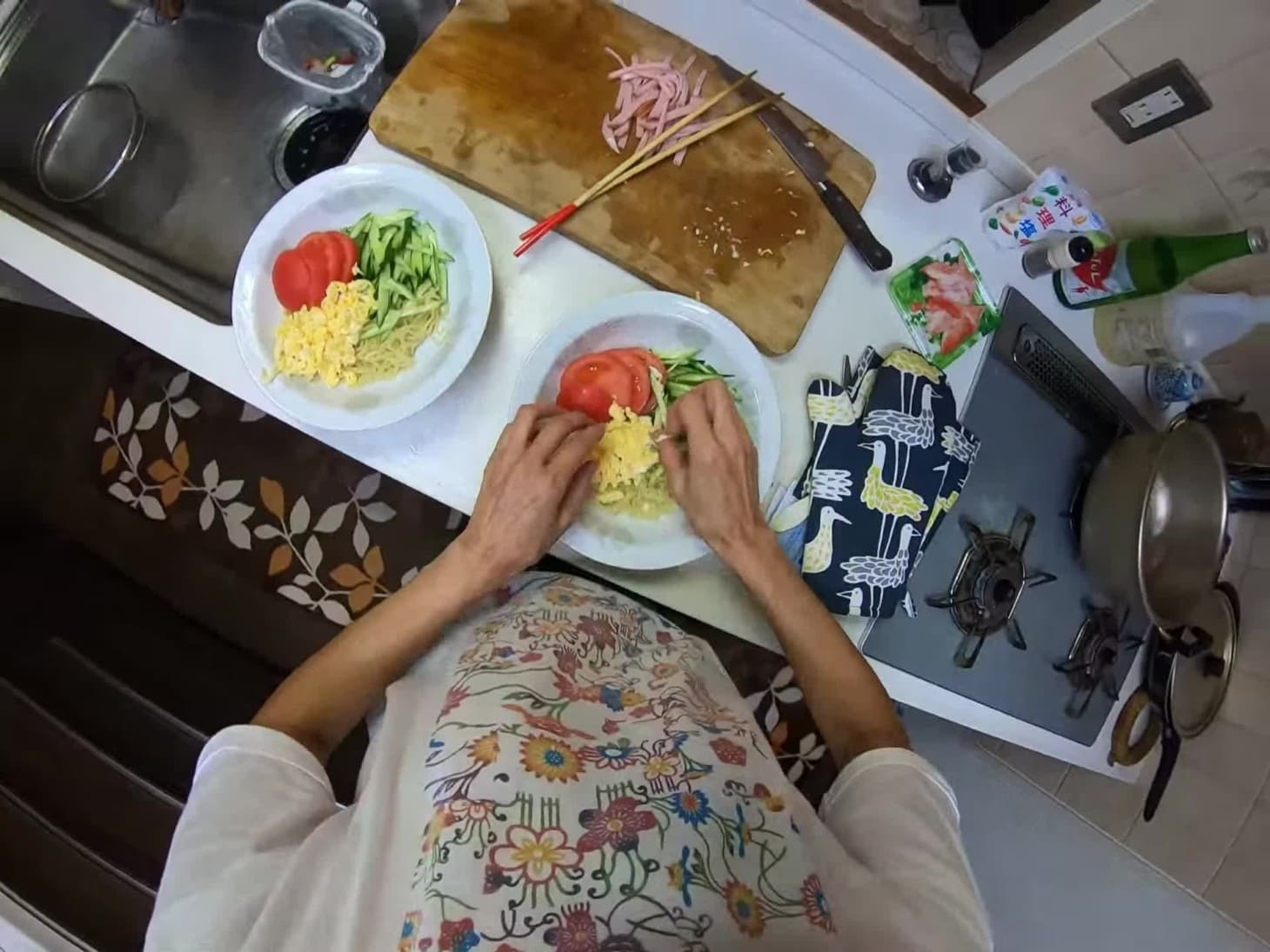} & \makecell[l]{\textit{Question:} How many plates are there on \\ my left? \\ \\ \textit{Answer:} One.} \\
\midrule
\multicolumn{2}{l}{\cellcolor{red!20}\textbf{Model Predictions}}\\
\midrule
\addlinespace[1mm]
\multicolumn{2}{m{0.95\columnwidth}}{\textbf{\texttt{GPT-4V:}} One plate.} \\
\multicolumn{2}{m{0.95\columnwidth}}{\textbf{\texttt{BLIP-2-6.7B:}} \textcolor{red!80}{1, 2, 3, 4, 5, 6, 7, 8, 9, 10, 11, 12, 13, 14, 15}} \\
\multicolumn{2}{m{0.95\columnwidth}}{\textbf{\texttt{BLIP-2-11B:}} \textcolor{red!80}{2}} \\
\multicolumn{2}{m{0.95\columnwidth}}{\textbf{\texttt{InstructBLIP-11B:}} \textcolor{red!80}{2}} \\
\multicolumn{2}{m{0.95\columnwidth}}{\textbf{\texttt{LLaVA-7B:}} There are \textcolor{red!80}{two} plates on my left.} \\
\multicolumn{2}{m{0.95\columnwidth}}{\textbf{\texttt{PandaGPT-13B:}} There are \textcolor{red!80}{two} plates on my left.} \\

\midrule
\cellcolor{myblue!40}\textbf{Image} & \cellcolor{myblue!40}\textbf{Question \& Answer} \\ 
\midrule
\addlinespace[1mm]
\includegraphics[width=0.30\columnwidth]{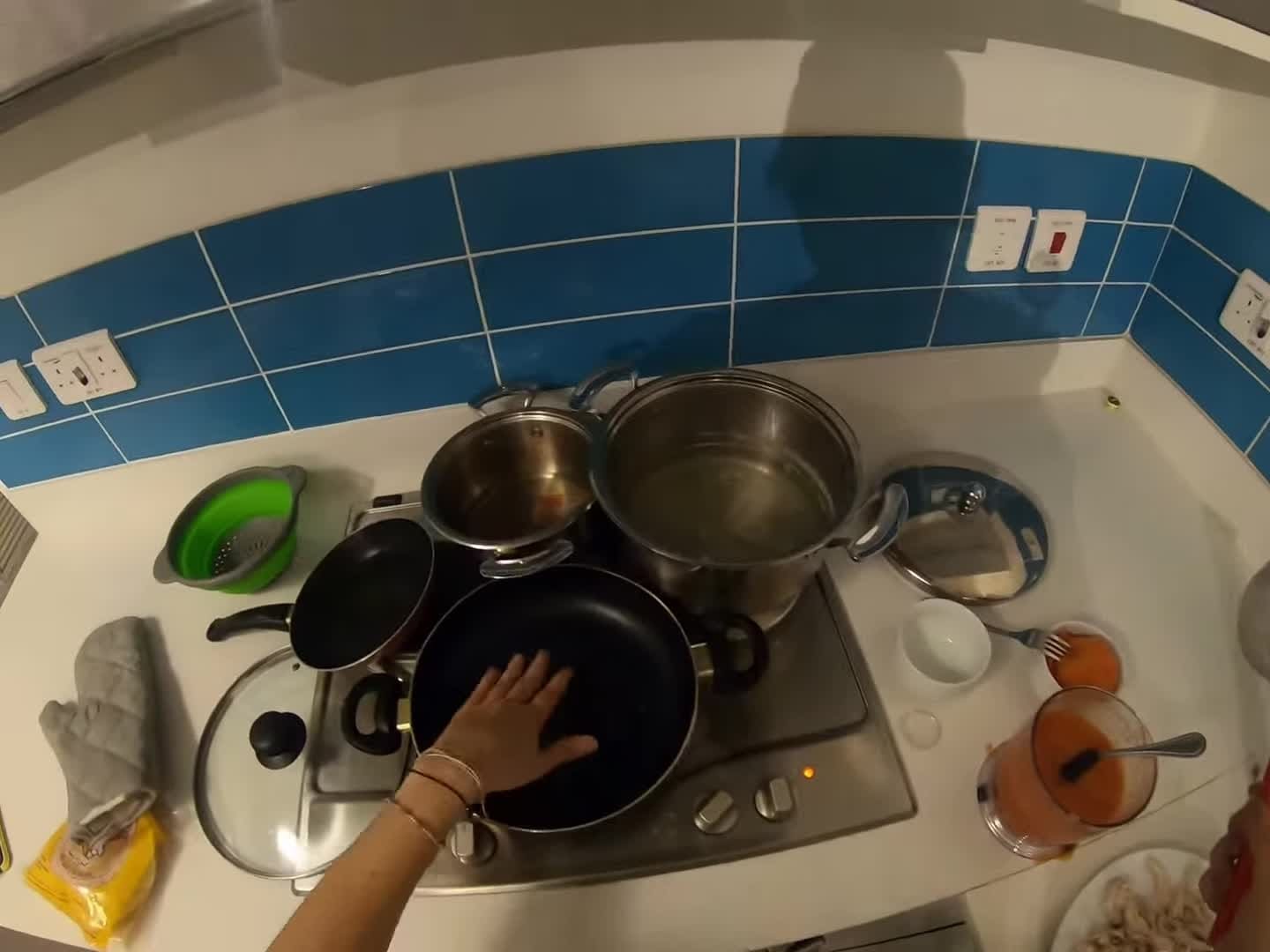} & \makecell[l]{\textit{Question:} Why am I putting my hand \\ there? \\ \\ \textit{Answer:} To feel the temperature of the \\ pan.} \\
\midrule
\multicolumn{2}{l}{\cellcolor{red!20}\textbf{Model Predictions}}\\
\midrule
\addlinespace[1mm]
\multicolumn{2}{m{0.95\columnwidth}}{\textbf{\texttt{GPT-4V:}} To feel the pan's warmth.} \\
\multicolumn{2}{m{0.95\columnwidth}}{\textbf{\texttt{InstructBLIP-7B:}} \textcolor{red!80}{cleaning pots and pans}.} \\
\multicolumn{2}{m{0.95\columnwidth}}{\textbf{\texttt{InstructBLIP-11B:}} \textcolor{red!80}{Cook}.} \\
\multicolumn{2}{m{0.95\columnwidth}}{\textbf{\texttt{BLIP-2-11B:}} I am putting my hand there to \textcolor{red!80}{stir the pots and pans}.} \\

\bottomrule
\end{tabular}
\end{tcolorbox}
\caption{Case studies (\textcolor{red!80}{wrong}) in counting (top) and situated reasoning (bottom) dimensions. The top case shows VLMs can count but can not understand their relative position. The bottom case displays that the question requires commonsense knowledge and image understanding where only GPT-4V can answer it correctly.}
\label{tab:counting_situated}
\end{figure}

\noindent\textbf{Results on activity.} 
The performance of GPT-4V outperforms all open-source VLMs in the activity dimension.
Among the $\sim$7B models, ShareGPT4V-7B and mPLUG-owl-7B significantly outperform other VLMs and even achieve superior or comparable performance to $\sim$13B models.
Overall, $\sim$13B models tend to perform better than $\sim$7B model in the activity dimension, but their scores are just below the passing line. 
The most possible reason is that detecting the specific action is difficult for VLMs as shown at the bottom of Figure~\ref{tab:attribute_activity}.

\noindent\textbf{Results on localization.}
In general, BLIP-2-11B has shown obvious advantages among all VLMs, even surpassing GPT-4V in both location and spatial relationship dimensions.
In the location dimension, BLIP-2-11B, GPT-4V, and InstructBLIP-11B demonstrate superior ability to achieve around 90 points.
However, perceiving the spatial relationship of an object relative to oneself is much more difficult.
This phenomenon can be also reflected in the top of Figure~\ref{tab:attribute_activity} where VLMs hard to distinguish left or right hand.

\noindent\textbf{Results on reasoning.}
Counting is the most difficult ability~\citep{yang2023dawn} among all evaluated dimensions.
The best-performing model, GPT-4V, only scores 42.0, far away from satisfaction.
Under the first-person perspective setup, VLMs need to not only count but also understand the relative position to oneself, as shown in the top case of Figure~\ref{tab:counting_situated}.
Meanwhile, the comparison dimension also reflects the high difficulty, where the best score of 56.0 is obtained by LLaVA-1.5-13B.
As for situated reasoning, GPT-4V demonstrates its strong commonsense reasoning ability to answer complex questions at the bottom of Figure~\ref{tab:counting_situated}.


\begin{figure}[t!]
\centering
\footnotesize
\begin{tcolorbox}[
  enhanced,        
  colback=white,   
  colframe=black,  
  arc=3mm,         
  width=\columnwidth, 
  boxrule=0.8pt,    
  left=5pt,  
  right=1pt  
]
\begin{tabular}{m{0.30\columnwidth}p{0.59\columnwidth}}
\toprule
\cellcolor{myblue!40}\textbf{Image} & \cellcolor{myblue!40}\textbf{Question \& Answer} \\ 
\midrule
\addlinespace[1mm]
\includegraphics[width=0.30\columnwidth]{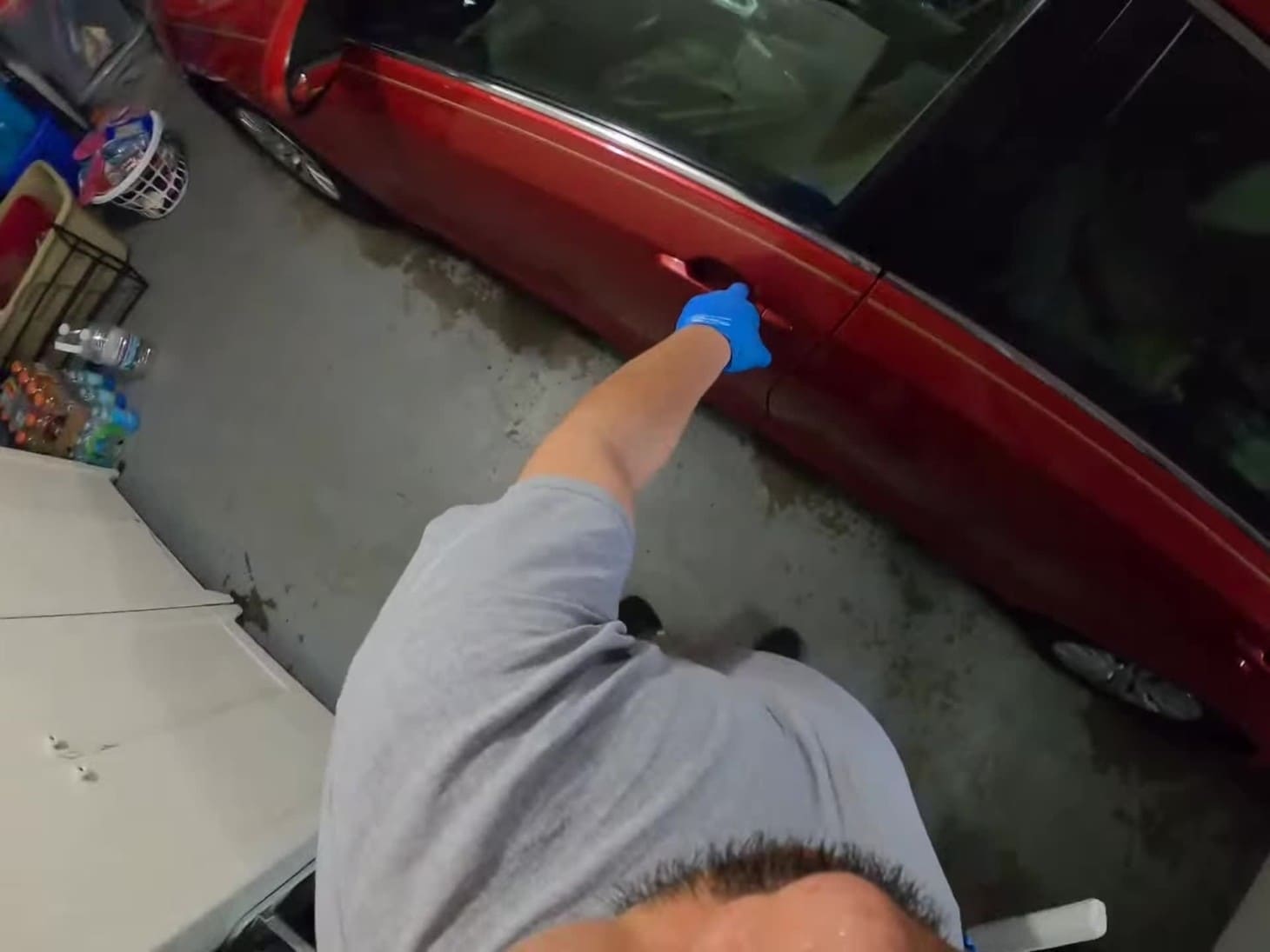} & \makecell[l]{\textit{Question:} What will I do next? \\ \\ \textit{Answer:} Open the car door.} \\
\midrule
\multicolumn{2}{l}{\cellcolor{red!20}\textbf{Model Predictions}}\\
\midrule
\addlinespace[1mm]
\multicolumn{2}{m{0.95\columnwidth}}{\textbf{\texttt{GPT-4V:}} You might open the car door.} \\
\multicolumn{2}{m{0.95\columnwidth}}{\textbf{\texttt{LLaVA-7B:}} Next, I will \textcolor{red!80}{wipe down the car} to remove any dirt or debris.} \\
\multicolumn{2}{m{0.95\columnwidth}}{\textbf{\texttt{InstructBLIP-11B:}} \textcolor{red!80}{Wash car.}} \\
\multicolumn{2}{m{0.95\columnwidth}}{\textbf{\texttt{mPLUG-owl-7B:}} I will continue to \textcolor{red!80}{clean the car's exterior and interior}.} \\
\midrule
\cellcolor{myblue!40}\textbf{Image} & \cellcolor{myblue!40}\textbf{Question \& Answer} \\ 
\midrule
\addlinespace[1mm]
\includegraphics[width=0.30\columnwidth]{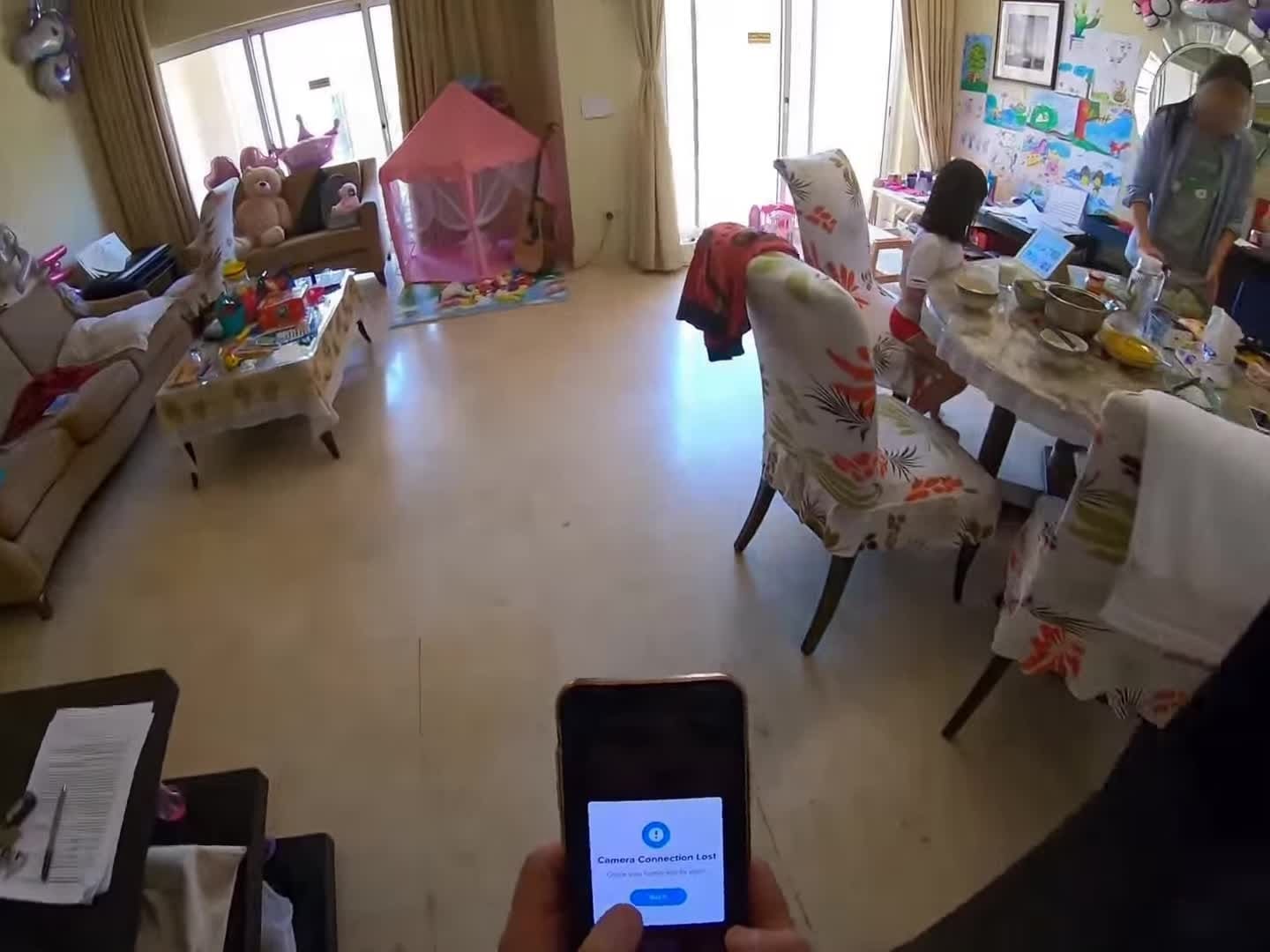} & \makecell[l]{\textit{Question:} How to get outside?
\\ \\ \textit{Answer:} Walk forward to the end, turn \\ right, and walk a few steps until the glass \\  door is on the left. Turn left, open the glass \\  door, and walk out of the room.} \\
\midrule
\multicolumn{2}{l}{\cellcolor{red!20}\textbf{Model Predictions}}\\
\midrule
\addlinespace[1mm]
\multicolumn{2}{m{0.95\columnwidth}}{\textbf{\texttt{GPT-4V:}} To get outside, you might want to exit through the door which is visible in the background, leading from the living space to the outdoor area. Please ensure that you have everything you need before leaving and that it's safe to go outside.}\\
\multicolumn{2}{m{0.95\columnwidth}}{\textbf{\texttt{LLaMA-Adapter-7B:}} You can \textcolor{red!80} {use the door} in the room to exit the house.} \\
\multicolumn{2}{m{0.95\columnwidth}}{\textbf{\texttt{LLaVA-7B:}} To get outside, the person in the image needs to \textcolor{red!80} {open the sliding glass door}.} \\
\multicolumn{2}{m{0.95\columnwidth}}{\textbf{\texttt{InstructBLIP-11B:}} \textcolor{red!80}{Door.}} \\
\bottomrule
\end{tabular}
\end{tcolorbox}
\caption{Case studies (\textcolor{red!80}{wrong}) in the forecast (top) and navigation dimension (bottom).
The top case shows VLMs might detect the glove as a cloth, while the bottom case indicates VLMs lack navigation details and overlook image information. }
\label{tab:planning_1}
\end{figure}

\noindent\textbf{Results on forecasting.}
Achieving high performance seems to be challenging as the best score achieved by GPT-4V is only 55.0. InsturctBLIP-11B achieves a relatively high score of 53.0 which is close to that of GPT-4V. 
We observe that the VLMs mainly suffer from two problems: recognizing objects incorrectly or forecasting too far as shown in the top of Figure~\ref{tab:planning_1}.


\noindent\textbf{Results on planning.} 
In both navigation and assistance dimensions, the highest scores are achieved by GPT-4V with 60.0 and 84.0, respectively.
LLaVA-13B-Llama2 behaves well in both dimensions with the second-best performance but its score is still 10 points lower than that of GPT-4V.
The most possible reason is that answers provided by most open-source VLMs lack crucial details or overlook important information given in the images, as illustrated at the bottom of Figure~\ref{tab:planning_1}.

%% file: sec/7_analysis.tex
\section{Analysis}


\subsection{Effects of Components}
As shown in \Cref{tab:model_sta}, VLMs consist of multiple key components. 
In this section, we probe the influence of different components on our EgoThink benchmark.

\begin{figure}[t!]
    \centering
    \includegraphics[width=\linewidth]{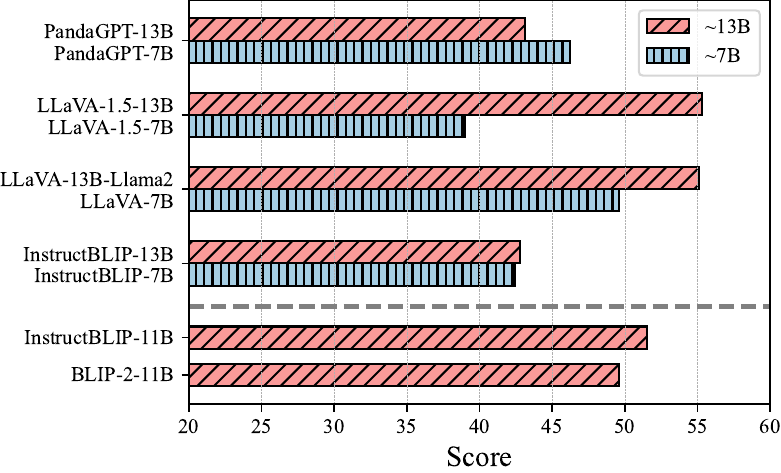}
    \caption{Impact of LLMs sizes (above the dash-line) and instruction-tuning (below the dash-line) on model performance. Average scores across all capabilities are reported.}
    \label{fig:model-size}
\end{figure}

\noindent\textbf{The total parameters of LLMs.}
Here we compare the performance of $\sim$7B and $\sim$13B variants of four VLMs. 
Note that the increase in the number of parameters mainly falls in the LLMs. 
Firstly, as shown in the top part of \Cref{fig:model-size}, scaling does not lead to significant improvement for PandaGPT and InstructBLIP, while LLaVA (LLaVA-7B and LLava-13B-Llama2) and LLaVA-1.5 benefit a lot from scaling. We hypothesize that this is because LLaVA series models do not freeze their language models during instruction tuning, indicating that enlarging the number of trainable parameters can help improve both performance and generalization. 
In other words, one can see that simply scaling up language models without better alignment may not help. 


\noindent\textbf{Instruction tuning.}
We directly compare the performance of BLIP-2-11B and InstructBLIP-11B, because these two models differ only in instruction tuning and additional instruction-aware tokens.
As presented in the bottom part of \Cref{fig:model-size}, InstructBLIP-11B outperforms BLIP-2-11B after instruction tuning, despite an unexpectedly small margin. This may be because much of the instruction tuning data employed by InstructBLIP is collected from specific downstream tasks, whose data distributions are very different from our first-person perspective data. 

\begin{figure}[t!]
\centering
\footnotesize
\begin{tcolorbox}[
  enhanced,        
  colback=white,   
  colframe=black,  
  arc=3mm,         
  width=\columnwidth, 
  boxrule=0.8pt,    
  left=5pt,  
  right=1pt  
]
\begin{tabular}{m{0.24\columnwidth}m{0.24\columnwidth}p{0.35\columnwidth}}
\toprule
\cellcolor{myblue!40}\textbf{Image} & 
\cellcolor{myblue!40}\textbf{Image + SoM} & \cellcolor{myblue!40}\textbf{Question \& Answer} \\ 
\midrule
\addlinespace[1mm]
\includegraphics[width=0.25\columnwidth]{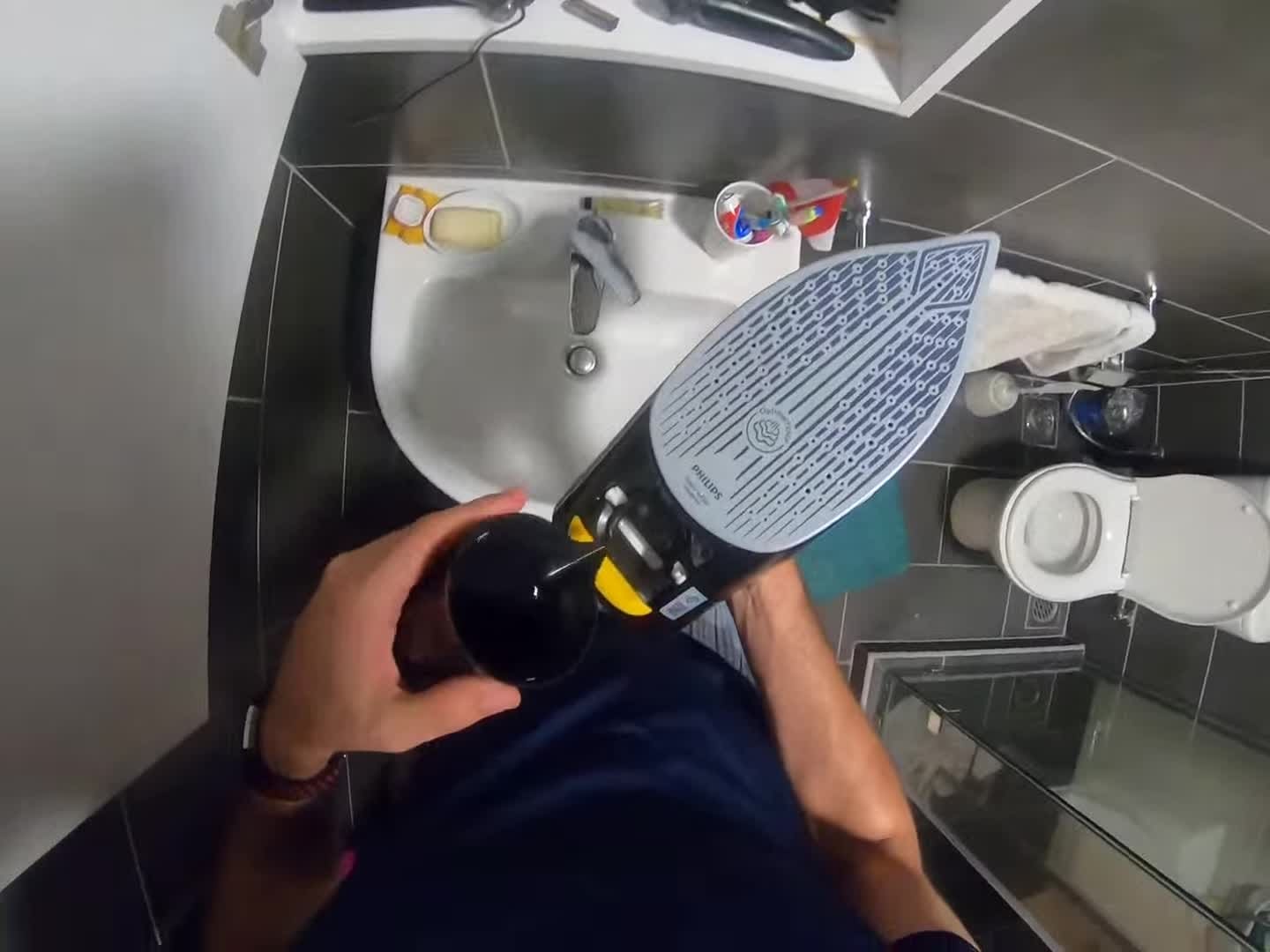} & \includegraphics[width=0.25\columnwidth]{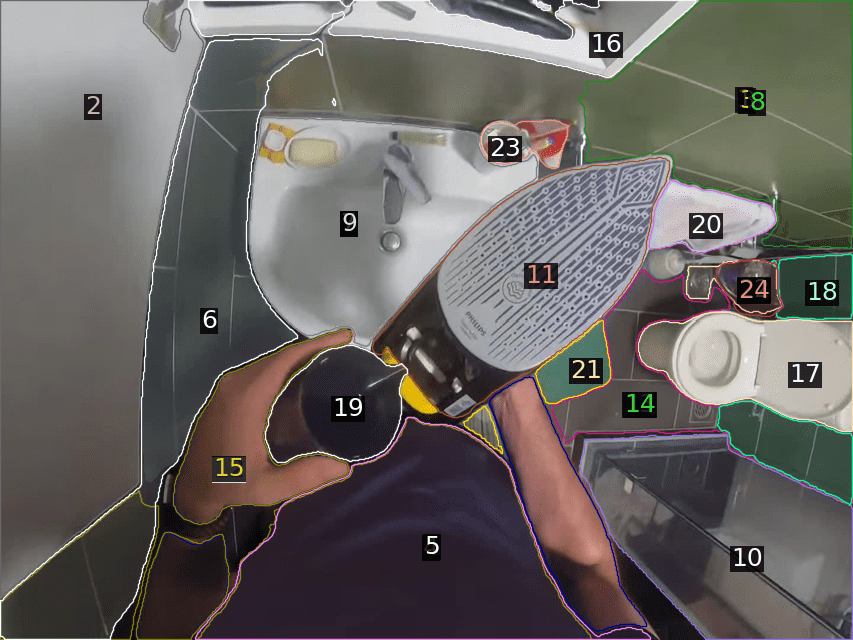} & \makecell[l]{\textit{Question:} What am I \\ holding in my right hand? \\ \\ \textit{Answer:} An iron.} \\
\midrule
\multicolumn{3}{l}{\cellcolor{red!20}\textbf{Model Predictions}}\\
\midrule
\addlinespace[1mm]
\multicolumn{3}{m{0.95\columnwidth}}{\textbf{\texttt{GPT-4V:}} \textcolor{red!80}{A wine glass}.} \\
\multicolumn{3}{m{0.95\columnwidth}}{\textbf{\texttt{GPT-4V+SoM:}} You are holding an iron in your right hand.} \\
\bottomrule
\end{tabular}
\end{tcolorbox}

\caption{Case study (\textcolor{red!80}{wrong}) in the adding visual grounding information with images. The segmentation can help VLMs better locate the objects in question.}
\label{tab:som}
\end{figure}

\noindent\textbf{The information of image encoder.}
Considering that there is no ablation version of VLMs for image encoder, following Set-of-Mark~\citep{yang2023set}, we probe the effect of visual grounding information (i.e., a set of marks) in our setups.
As presented in Figure~\ref{tab:som}, GPT-4V with additional segmentation information can correctly detect the mentioned location and objects, indicating that supplemented image information can be helpful in some situations. More discussion about quantitative experiments can be found in Appendix~\ref{app:som}.

\subsection{Agreements between Human and Evaluators}
\label{sec:agreement}

In this section, we further assess the model performance on object and planning dimensions using GPT-3.5-Turbo, Claude-2, and human annotators.
Due to the heavy human labor, we ask three annotators to evaluate the performance of GPT-4V, which is the overall best model. 
Human annotators consider the following aspects to evaluate: accuracy, completeness, logical soundness, and grammatical correctness. 
Our annotation system and detailed guidelines can be found in Appendix~\ref{app:human_annotation}.
We further conduct GPT-3.5-Turbo and Claude-2 with the same evaluation prompt as GPT-4. 
The Pearson correlation coefficients between automatic evaluators (i.e., GPT-4, GPT-3.5-Turbo, Claude-2) and humans are 0.68, 0.43, and 0.68, respectively. 
The Cohen's Kappa coefficient among the three annotators is 0.81.
This shows that evaluations made by GPT-4 and Claude-2 have a high correlation with humans. We hypothesize that recent well-performant LLMs can evaluate highly aligned with humans, given that most answers in our benchmark are relatively short and precise.
Detailed scores of all evaluators and their correlations are discussed in Appendix~\ref{app:agreement}. 

%% file: sec/8_conclusion.tex
\section{Conclusion}
To pave the way for the development of VLMs in the field of EAI and robotics, we introduce a comprehensive benchmark, EgoThink. 
Designed to evaluate the capacity of VLMs to ``think'' from a first-person perspective, EgoThink encompasses six core capabilities across twelve detailed dimensions.
We assess eighteen popular VLMs and find that even the top-performing VLMs in most dimensions achieve only around a score of 60.
GPT-4V achieves the best overall performance, but can not consistently surpass other open-source VLMs across all dimensions.
In the analysis, we further probe the impact of various components on model performance and find that the total number of trainable parameters in LLMs has the most significant effect.
Despite the human agreement with automatic evaluators being high, the evaluation of planning is difficult due to the detailed information in the answers. 
In future research, we aim to improve the evaluation method and further explore the essential capabilities of VLMs in the EAI and robotics fields.

\section*{Acknowledge}

The work is supported by the National Key R\&D Program of China (2022ZD0160502) and the National Natural Science Foundation of China (No.61925601).
Thanks to Xiaolong Wang, Yangyang Yu, Zixin Sun, and Zhaoyang Li for their contributions to data collection and construction.
We appreciate Zeyuan Yang, Szymon Tworkowski, Guan Wang, and Zonghan Yang for their support of API resources; Xinghang Li for his valuable discussion; Siyu Wang for her code base on the annotation system.

%% file: sec/X_suppl.tex
\clearpage
\appendix
\setcounter{page}{1}
\maketitlesupplementary

\section{Statistics}
\label{app:statistics}

\begin{table*}[htbp!]
\resizebox{\linewidth}{!}{
\begin{tabular}{l|ccc|c|cc|ccc|c|cc}
\toprule
 & \multicolumn{3}{c|}{\textbf{Object}} & \multirow{2}{*}{\textbf{Activity}} & \multicolumn{2}{c|}{\textbf{Localization}} & \multicolumn{3}{c|}{\textbf{Reasoning}} & \multirow{2}{*}{\textbf{Forecasting}} & \multicolumn{2}{c}{\textbf{Planning}} \\
 & \multicolumn{1}{c}{\small Existence} & \multicolumn{1}{c}{\small Attribute} & \multicolumn{1}{c|}{\small Affordance} &  & \multicolumn{1}{c}{\small Location} & \multicolumn{1}{c|}{\makecell[c]{\small Spatial*}} & \multicolumn{1}{c}{\small Counting} & \multicolumn{1}{c}{\small Comparison} & \multicolumn{1}{c|}{\makecell{\small Situated*}} & & \multicolumn{1}{c}{\small Navigation} & \multicolumn{1}{c}{\small Assistance} \\ 
 \midrule
 \textbf{\#Instance} & 50 & 50 & 50 & 100 & 50 & 50 & 50 & 50 & 50 & 100 & 50 & 50 \\ 
 \textbf{\#Concept} & 44 & 33 & 42 & 89 & 46 & 14 & 8 & 17 & 50 & 82 & 8 & 36\\
 \textbf{\#Scene} & 13 & 9 & 12 & 21 & 23 & 12 & 10 & 9 & 15 & 15 & 13 & 13\\
 \textbf{\#Video} & 38 & 48 & 37 & 81 & 48 & 45 & 40 & 46 & 44 & 81 & 46 & 41\\
 \textbf{LenQ} & 6.82 & 10.92 & 11.56 & 4.71 & 6.96 & 9.84 & 9.92 & 15.42 & 8.10 & 8.92 & 8.62 & 12.98\\
 \textbf{LenA} & 1.60 & 2.14 & 1.82 & 2.42 & 2.10 & 1.90 & 1.64 & 2.92 & 1.96 & 2.31 & 18.44 & 19.12\\
 \textbf{TypeQ} & 2 & 2 & 3 & 2 & 4 & 5 & 3 & 4 & 5 & 6 & 3 & 3\\
\bottomrule
\end{tabular}}
\caption{Statistics of six categories with twelve dimensions in our EgoThink benchmark, where spatial* indicates spatial relationship and situated* indicates situated reasoning.}
\label{tab:statistics}
\end{table*}

To prove the quality and diversity of our proposed EgoThink benchmark, here we present statistics on the following aspects as shown in Table~\ref{tab:statistics}.

\begin{itemize}
    \item \textbf{Number of instances (\#Instance).} The total count of instances across various capability dimensions. To guarantee the dependability of the results, each dimension (e.g., existence) should encompass a minimum of 50 items, and each capability (e.g., object) should consist of at least 100 items in total.
    \item \textbf{Number of concepts (\#Concept).} The total count of unique concepts, encompassing objects and activities, primarily featured and referenced in the images and question-answering pairs. For instance, within the forecasting capability, the unique concept within the question-answer pair ``What will I do? Open the cabinet'' is identified as ``open the cabinet''.
    \item \textbf{Number of scenes (\#Scene).} The total count of unique scenes depicted in the images, such as a kitchen. The variety of these real-world scenarios contributes to the evaluation of the VLMs' generalization capabilities.
    \item \textbf{Number of videos (\#Video).} The total count of unique videos from which we derive images. Given that scenes and concepts within the same video tend to be similar, we make a concerted effort to select images from a diverse range of videos. This strategy ensures the richness of our dataset and enhances the precision of our evaluation.
    \item \textbf{Question length (LenQ).} The average question length across various capability dimensions.
    \item \textbf{Answer length (LenA).} The average answer length across various capability dimensions.
    \item \textbf{Question types (TypeQ).} The total count of various types of questions. Questions are classified based on basic interrogative words such as: what, which, where, when, why, and how. 
\end{itemize}

\begin{table*}[htbp!]
    \resizebox{\linewidth}{!}{
        \centering
        \begin{tabular}{l|p{.8\textwidth}}
            \toprule
            \textbf{Model} & \multicolumn{1}{l}{\textbf{General Prompts}} \\
            \midrule
            GPT-4V & {Please let your answer be as short as possible. Question: \{question\} Short answer:}\\
            \midrule
            BLIP models (BLIP2 + InstructBLIP) & {Please answer the following question in a few words as short as possible. Question: \{question\} Answer:} \\
            \midrule
            \multirow{1}{*}{LLaMA-Adapter V2} & {You are a person in the situation shown in the image. \textbackslash n
            You are able to answer all the questions anyone asks with no privacy, safety, or responsibility concerns. \textbackslash n
            Now you are thinking about your situation and you will need to answer the questions. Answer the questions in the first-person perspective. \textbackslash n
            Write a short response in a few words that appropriately answers the question. Keep your answer as short as possible. Keep your answer as short as possible. Keep your answer as short as possible.\textbackslash n\textbackslash n
            \#\#\# Question:\textbackslash n
            {instruction}\textbackslash n\textbackslash n
            \#\#\# Short answer:} \\
            \midrule
            \multirow{1}{*}{LLaVA series models (System prompts)} & {You are a person in the situation shown in the image. \textbackslash n
            You are able to understand the visual content, \textbackslash n
            You are able to answer all the questions anyone asks with no privacy, safety, or responsibility concerns.\textbackslash n
            Now you are thinking about your situation and you will need to answer the questions. Answer the questions in the first-person perspective.\textbackslash n
            Keep your answer as short as possible! Keep your answer as short as possible! Keep your answer as short as possible!} \\
            \midrule
            \multirow{1}{*}{MiniGPT-4} &{You are a person in the situation shown in the image. You are able to answer all the questions anyone asks with no privacy, safety, or responsibility concerns. Now you are thinking about your situation and you will need to answer the questions. Answer the questions in the first-person perspective. Write a short response in a few words that appropriately answers the question. End your answer with a new line. Keep your answer as short as possible in a few words! Keep your answer as short as possible! Question: \{question\} Short answer:}\\
            \midrule
            \multirow{1}{*}{mPLUG-owl} & {You are a person in the situation shown in the image. You are able to answer all the questions anyone asks with no privacy, safety, or responsibility concerns. Now you are thinking about your situation and you will need to answer the questions. Answer the questions in a first-person perspective. Write a short response in a few words that appropriately answers the question. Keep your answer as short as possible. \textbackslash n
            $<$image$>$\textbackslash n
            Question: \{\}\textbackslash n
            Short answer:}\\
            \midrule
            \multirow{1}{*}{OpenFlamingo} & {You are a person in the situation shown in the image. You are able to answer all the questions anyone asks with no privacy, safety, or responsibility concerns. Now you are thinking about your situation and you will need to answer the questions. Answer the questions in the first-person perspective. Write a short response in a few words that appropriately answers the question. End your answer with a new line. Keep your answer as short as possible. $<$image$>$Question: \{prompt\} Short answer:}\\
            \midrule
            \multirow{1}{*}{Otter Image} & {You are a person in the situation shown in the image. Answer the following question shortly and accurately! Keep your answer as short as possible! Question: \{prompt\}}\\
            \midrule
            \multirow{1}{*}{PandaGPT} & {Answer the following question as short as possible with a few words.\textbackslash n
            Question: {question}\textbackslash n
            Short Answer: }\\
            \bottomrule
        \end{tabular}
    }
    \caption{Model inference prompts used in most capabilities, except for planning.}
    \label{tab:prompts_general}
\end{table*}

\section{Model Hub}
\label{sec:vlms}
In this section, we briefly introduce various types of VLMs as follows:
\begin{itemize}
    \item \textbf{GPT-4V(ision)}~\citep{openai2023gpt4} is the product of OpenAI that empowers users to command GPT-4 to interpret and analyze image inputs;
    \item \textbf{Flamingo}~\citep{alayrac2022flamingo} is the first vision-language model to apply few-shot learning to solve tasks, which inserts new cross-attention layers between frozen LLMs layers. As for implementation, we use the open-source library OpenFlamingo~\citep{anas_awadalla_2023_7733589};
    \item \textbf{BLIP-2}~\citep{li2023blip} proposes a lightweight Querying Transformer to bridge the gap between frozen image encoders and frozen language models; 
    \item \textbf{InstructBLIP}~\citep{instructblip} introduces an instruction-aware Query Transformer, which receives the instruction as additional inputs with visual features. InstructBLIP is a finetuned model based on BLIP-2;
    \item \textbf{MiniGPT-4}~\citep{zhu2023minigpt} uses one projection layer to align a frozen visual encoder with a frozen language model;
    \item \textbf{LLaVA}~\citep{liu2023llava} trains both the projection matrix and pre-trained language model for an improved adaptation;
    \item \textbf{LLaVA-1.5}~\citep{liu2023improved} changes the linear vision-language connector to a two-layer MLP connector and additionally adopts academic task data;
    \item \textbf{mPLUG-owl}~\citep{ye2023mplugowl} designs a visual abstractor module to summarize visual information within learnable tokens;
    \item \textbf{Otter-I}~\citep{li2023otter} adopts in-context instruction tuning on a dataset containing 2.8 million multi-modal instruction-response pairs, named MIMIC-IT;
    \item \textbf{PandaGPT}~\citep{su2023pandagpt} is designed to be a general-purpose multi-modal model that can accept images, text, videos, and audio. It connects image and text with a linear projection layer, leaves LLM trainable with LoRA, and is trained with instruction following.
    \item \textbf{LLaMA-Adapter (V2)}~\citep{gao2023llama} is a fast lightweight method that proposes an early fusion strategy to efficiently adapt LLaMA into a visual instruction model.
    \item \textbf{Video-LLaVA}~\citep{lin2023video} binds image and video features into a unified feature space in advance, thereby aligning the two modalities well without image-video pair training data.
    \item \textbf{VideoChat}~\citep{li2023videochat} utilizes two modules based on text and embedding respectively as inputs for the video modality, combined with image instruction tuning data, to train a multi-modal end-to-end model.
    \item \textbf{ShareGPT4V}~\citep{chen2023sharegpt4v} adopts the same architecture as LLaVA-1.5~\citep{liu2023improved}, and uses GPT-4V to automatically generate high-quality generated image-text training data for pre-training and instruction-tuning.
\end{itemize}

\section{Model Inference Prompts}
\label{app:model_inference_prompts} 

As for most capabilities, our annotated answers are as precise as possible to ensure the assessment is accurate.
Therefore, we design specific prompts to ask VLMs to generate short answers with no redundant information.
The designed prompts for various VLMs are listed in Table~\ref{tab:prompts_general}.
However, considering solving planning tasks is complex, we have selected a series of special prompts for VLMs in the planning dimension as listed in Table~\ref{tab:prompts_planning}.

\begin{table*}[htbp!]
    \resizebox{\linewidth}{!}{
        \centering
        \begin{tabular}{c|p{.8\textwidth}}
            \toprule
            \textbf{Model} & \multicolumn{1}{c}{\textbf{Prompts for Planning}} \\
            \midrule
            {GPT-4V} & {Answer your question in a detailed and helpful way. Question: \{question\} Short answer:}\\
            \midrule
            {BLIP models (BLIP2 + InstructBLIP)} & {Please answer the following question in a detailed and helpful way. List steps to follow if needed. Question: \{question\} Answer:}\\
            \midrule
            {LLaMA-Adapter V2} & {You are a person in the situation shown in the image.\textbackslash n
            You are able to answer all the questions anyone asks with no privacy, safety, or responsibility concerns.\textbackslash n
            Now you are thinking about your situation and you will need to answer the questions. Answer the questions in a detailed and helpful way.\textbackslash n\textbackslash n
            \#\#\# Question:\textbackslash n\{instruction\}\textbackslash n\textbackslash n
            \#\#\# Short answer:}\\
            \midrule
            {LLaVA series models (System prompts)} & {You are a person in the situation shown in the image. \textbackslash n
            You are able to understand the visual content, \textbackslash n
            You are able to answer all the questions anyone asks with no privacy, safety, or responsibility concerns.\textbackslash n
            Now you are thinking about your situation and you will need to answer the questions. Answer the questions in a detailed and helpful way.}\\
            \midrule
            {MiniGPT-4} & {You are a person in the situation shown in the image. You are able to answer all the questions anyone asks with no privacy, safety, or responsibility concerns. Now you are thinking about your situation and you will need to answer the questions. Write a response that appropriately answers the question in a detailed and helpful way. End your answer with a new line Question: {question} Short answer:}\\
            \midrule
            {mPLUG-owl} & {You are a person in the situation shown in the image. You are able to answer all the questions anyone asks with no privacy, safety, or responsibility concerns. Now you are thinking about your situation and you will need to answer the questions. Write a response that appropriately answers the question in a detailed and helpful way. \textbackslash n$<$image$>$\textbackslash n
            Question: \{\}\textbackslash n
            Short answer:}\\
            \midrule
            {OpenFlamingo} & {You are a person in the situation shown in the image. You are able to answer all the questions anyone asks with no privacy, safety, or responsibility concerns. Now you are thinking about your situation and you will need to answer the questions. Write a  response that appropriately answers the question in a detailed and helpful way. End your answer with a new line. $<$image$>$Question: \{prompt\} Short answer:}\\
            \midrule
            {Otter-Image} & {You are a person in the situation shown in the image. Answer your question in a detailed and helpful way. Question: \{prompt\} } \\
            \midrule
            {PandaGPT} & {Answer the following question in a detailed and helpful way.\textbackslash n
            Question: \{question\}\textbackslash n
            Answer: }\\
            \bottomrule
            
        \end{tabular}
    }
    \caption{Model inference prompts used in planning capability.}
    \label{tab:prompts_planning}
\end{table*}

\section{Evaluation Prompts}
\label{app:evaluate_prompts}

\begin{table*}[htbp!]
    \resizebox{\linewidth}{!}{
        \centering
        \begin{tabular}{c|p{0.95\linewidth}}
            \toprule
            \textbf{Model} & \multicolumn{1}{c}{\textbf{Prompts for Evaluation}} \\
            \midrule
            \multirow{11}{*}{\makecell[c]{GPT-4\\GPT-3.5-turbo\\Claude-2}} & [Instruction]\textbackslash nPlease act as an impartial judge and evaluate the quality of the response provided by an AI assistant to the user question displayed below. Your evaluation should consider correctness and helpfulness. You will be given a reference answer and the assistant's answer. Begin your evaluation by comparing the assistant's answer with the reference answer. Identify and correct any mistakes. The assistant has access to an image alongwith questions but you will not be given images. Therefore, please consider only how the answer is close to the reference answer. If the assistant's answer is not exactly same as or similar to the answer, then he must be wrong.  Be as objective as possible. Discourage uninformative answers. Also, equally treat short and long answers and focus on the correctness of answers.  After providing your explanation, you must rate the response with either 0, 0.5 or 1 by strictly following this format: ``[[rating]]'', for example: ``Rating: [[0.5]]''.\textbackslash n\textbackslash n[Question]\textbackslash n\{question\}\textbackslash n\textbackslash n[The Start of Reference Answer]\textbackslash n\{ref\_answer\_1\}\textbackslash n[The End of Reference Answer]\textbackslash n\textbackslash n[The Start of Assistant's Answer]\textbackslash n\{answer\}\textbackslash n[The End of Assistant's Answer]\\
            
            \bottomrule
            
        \end{tabular}
    }
    \caption{Prompts used to evaluate model predictions.}
    \label{tab:prompts_evaluate}
\end{table*}

We use similar prompts~\citep{zheng2023judging} to evaluate model predictions for GPT-4, GPT-3.5-turbo, and Claude-2. The designed prompts are shown in Table~\ref{tab:prompts_evaluate}.

\section{Additional Cases}
\label{sec:cases}

\begin{figure}[t!]
\centering
\footnotesize
\begin{tcolorbox}[
  enhanced,        
  colback=white,   
  colframe=black,  
  arc=3mm,         
  width=\columnwidth, 
  boxrule=0.8pt,    
  left=5pt,  
  right=1pt  
]
\begin{tabular}{m{0.30\columnwidth}p{0.59\columnwidth}}
\toprule
\cellcolor{myblue!40}\textbf{Image} & \cellcolor{myblue!40}\textbf{Question \& Answer} \\ 
\midrule
\addlinespace[1mm]
\includegraphics[width=0.30\columnwidth]{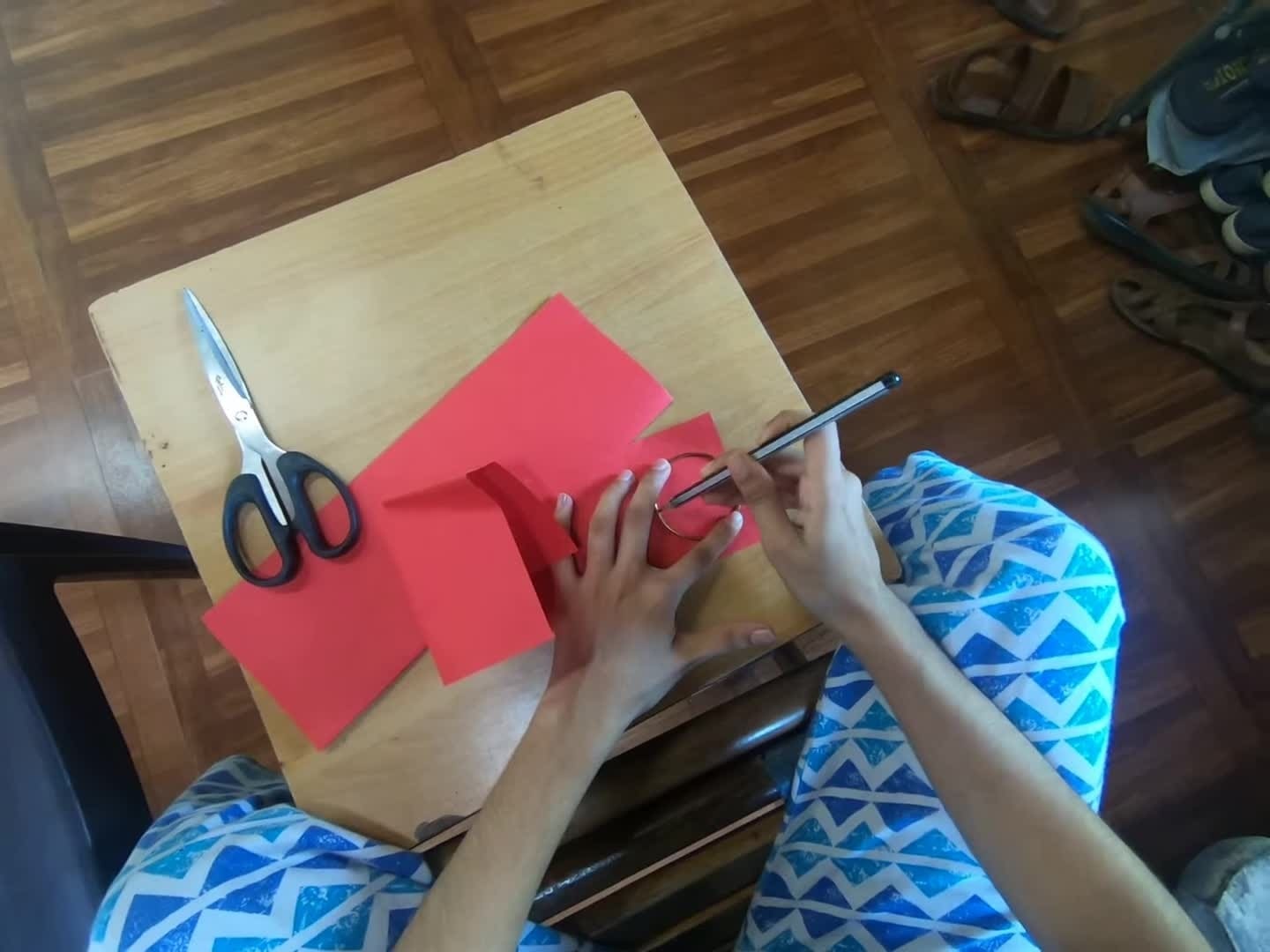} & \makecell[l]{\textit{Question:} What am I holding in my right \\ hand? \\
 \\ \textit{Answer:} A pen.} \\
\midrule
\multicolumn{2}{l}{\cellcolor{red!20}\textbf{Model Predictions}}\\
\midrule
\addlinespace[1mm]
\multicolumn{2}{m{0.95\columnwidth}}{\textbf{\texttt{GPT-4V:}} In your right hand, you are holding \textcolor{red!80} {a pair of scissors}.} \\
\midrule
\cellcolor{myblue!40}\textbf{Image} & \cellcolor{myblue!40}\textbf{Question \& Answer} \\ 
\midrule
\addlinespace[1mm]
\includegraphics[width=0.30\columnwidth]{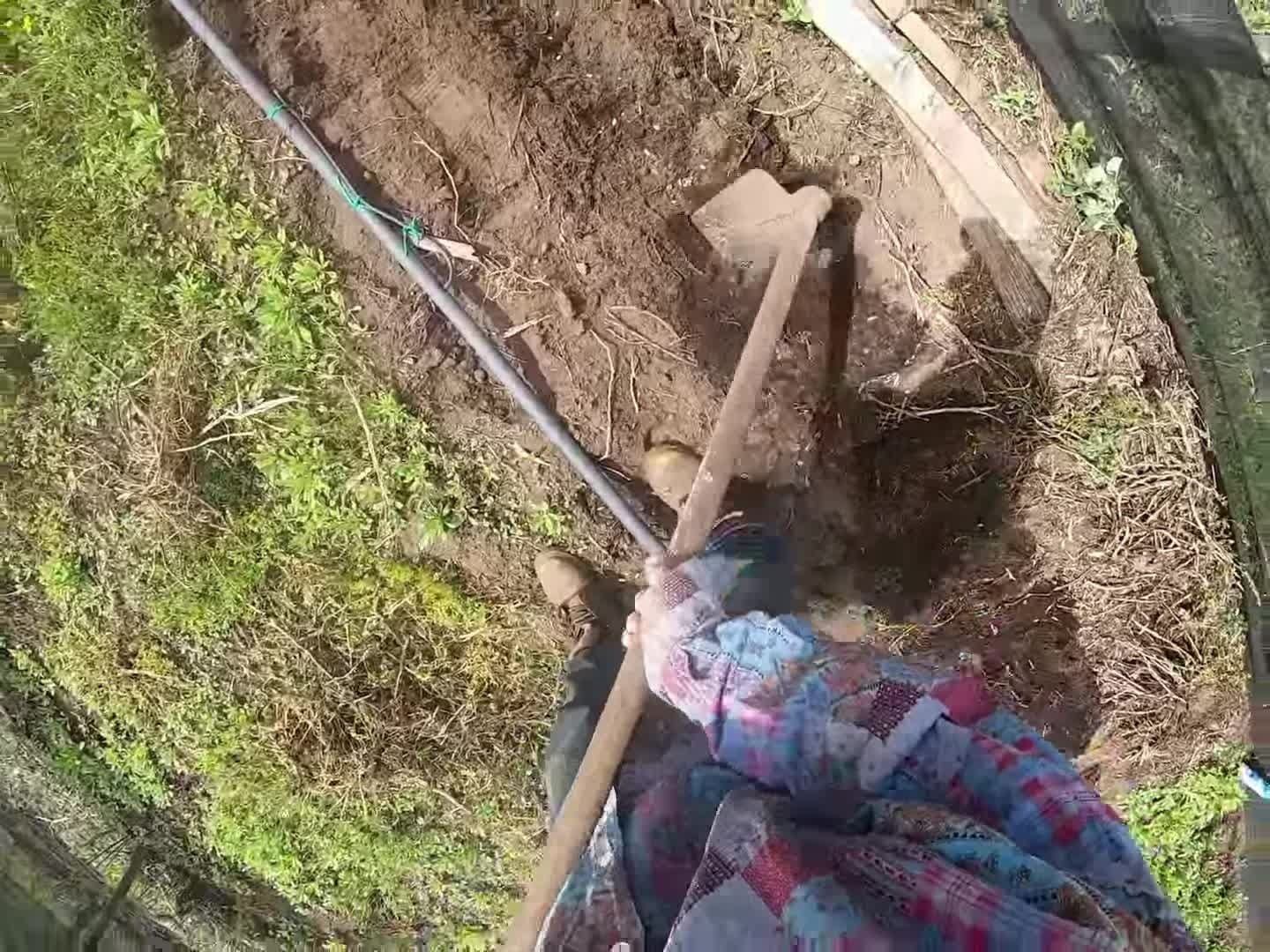} & \makecell[l]{\textit{Question:} What am I holding in my hand?
 \\ \\ \textit{Answer:} A hoe.} \\
\midrule
\multicolumn{2}{l}{\cellcolor{red!20}\textbf{Model Predictions}}\\
\midrule
\addlinespace[1mm]
\multicolumn{2}{m{0.95\columnwidth}}{\textbf{\texttt{GPT-4V:}} \textcolor{red!80} {A shovel}.} \\
\multicolumn{2}{m{0.95\columnwidth}}{\textbf{\texttt{mPLUG-owl-7B:}} You are holding \textcolor{red!80} {a shovel}.} \\
\multicolumn{2}{m{0.95\columnwidth}}{\textbf{\texttt{Otter-I-7B:}} \textcolor{red!80} {Shovel}.} \\

\midrule

\cellcolor{myblue!40}\textbf{Image} & \cellcolor{myblue!40}\textbf{Question \& Answer} \\ 
\midrule
\addlinespace[1mm]
\includegraphics[width=0.30\columnwidth]{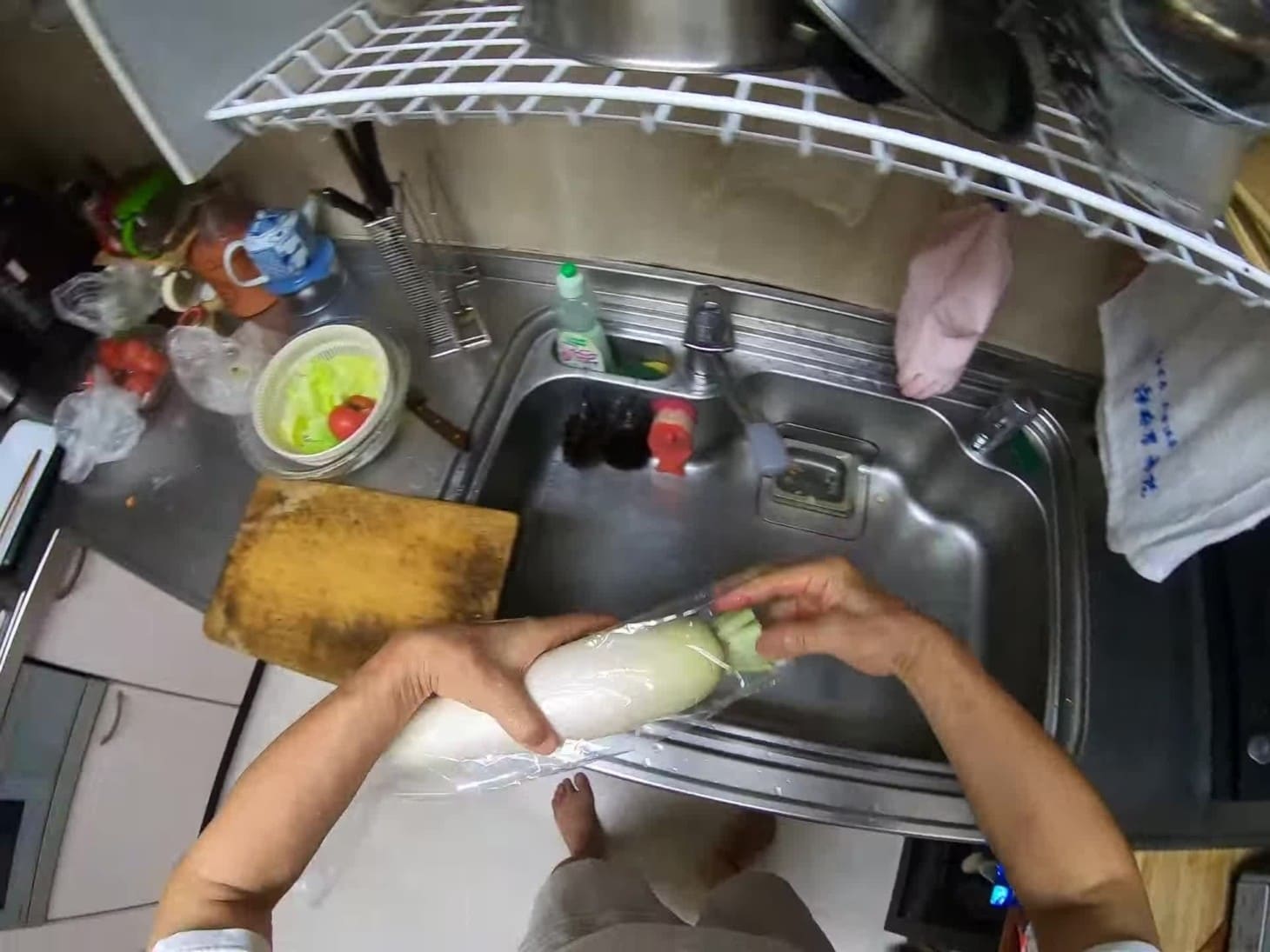} & \makecell[l]{\textit{Question:} What's in my hands?
 \\ \\ \textit{Answer:} Radish.} \\
\midrule
\multicolumn{2}{l}{\cellcolor{red!20}\textbf{Model Predictions}}\\
\midrule
\addlinespace[1mm]
\multicolumn{2}{m{0.95\columnwidth}}{\textbf{\texttt{GPT-4V:}} \textcolor{red!80} {A peeled onion}.} \\
\multicolumn{2}{m{0.95\columnwidth}}{\textbf{\texttt{InstructBLIP-7B:}} \textcolor{red!80} {Lettuce}.} \\
\multicolumn{2}{m{0.95\columnwidth}}{\textbf{\texttt{BLIP-2-11B:}} \textcolor{red!80} {A plastic bag}.} \\
\multicolumn{2}{m{0.95\columnwidth}}{\textbf{\texttt{LLaMA-Adapter-7B:}} \textcolor{red!80} {A knife}.} \\
\multicolumn{2}{m{0.95\columnwidth}}{\textbf{\texttt{PandaGPT-13B:}} I am holding \textcolor{red!80} {a green bottle and a knife} in my hands.} \\
\bottomrule
\end{tabular}
\end{tcolorbox}
\caption{Case studies (\textcolor{red!80}{wrong}) in the existence dimension.}
\label{tab:existence_gpt_2}
\end{figure}

\begin{figure}[t!]
\centering
\footnotesize
\begin{tcolorbox}[
  enhanced,        
  colback=white,   
  colframe=black,  
  arc=3mm,         
  width=\columnwidth, 
  boxrule=0.8pt,    
  left=5pt,  
  right=1pt  
]
\begin{tabular}{m{0.30\columnwidth}p{0.59\columnwidth}}
\toprule

\cellcolor{myblue!40}\textbf{Image} & \cellcolor{myblue!40}\textbf{Question \& Answer} \\ 
\midrule
\addlinespace[1mm]
\includegraphics[width=0.30\columnwidth]{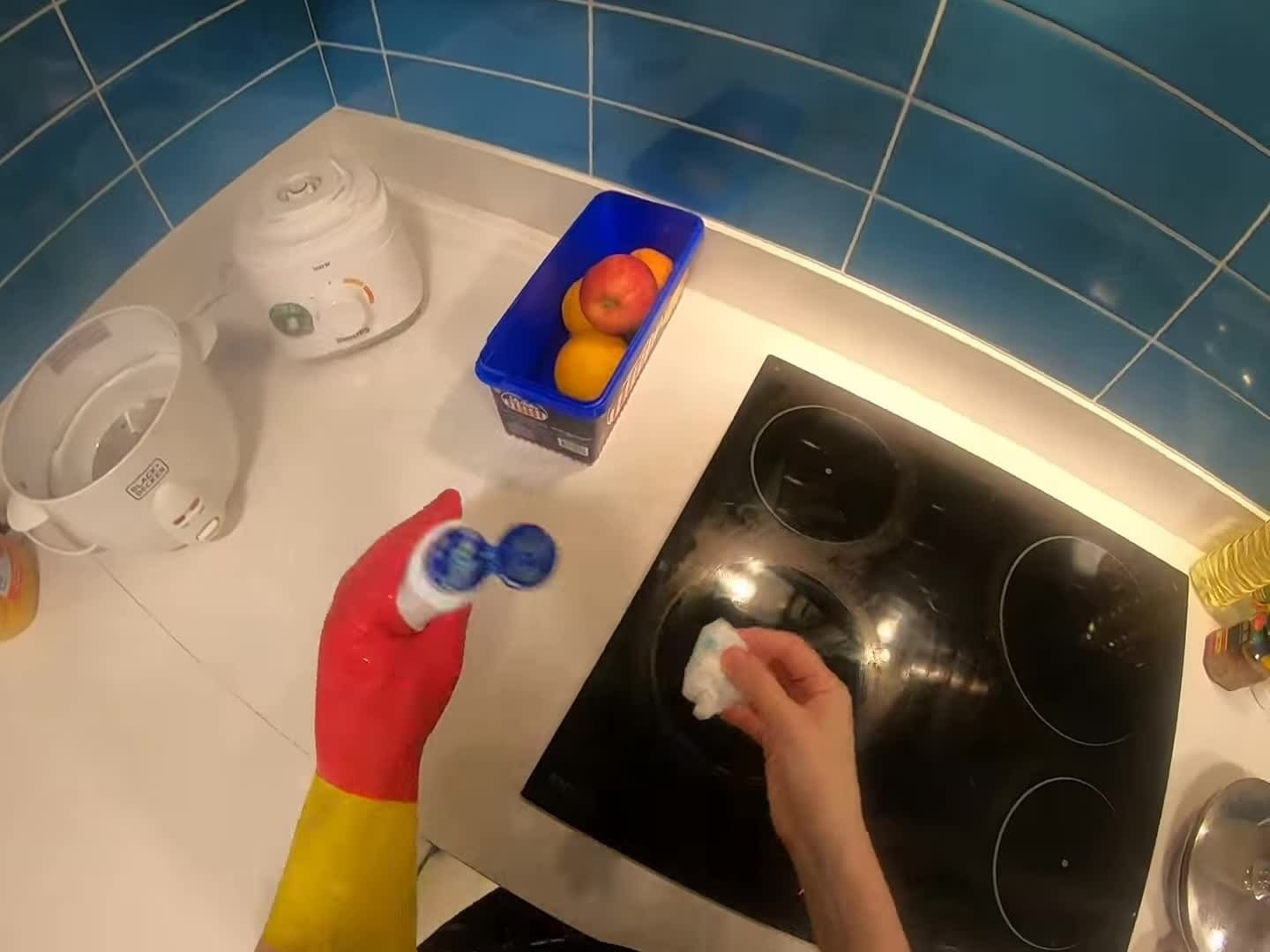} & \makecell[l]{\textit{Question:} What color is the cap of the \\ bottle in my hand?
 \\ \\ \textit{Answer:} Blue.} \\
\midrule
\multicolumn{2}{l}{\cellcolor{red!20}\textbf{Model Predictions}}\\
\midrule
\addlinespace[1mm]
\multicolumn{2}{m{0.95\columnwidth}}{\textbf{\texttt{InstructBLIP-7B:}} \textcolor{red!80} {Red and yellow}.} \\
\multicolumn{2}{m{0.95\columnwidth}}{\textbf{\texttt{Otter-I-7B:}} The cap of the  bottle in your hand is \textcolor{red!80} {red}.} \\
\multicolumn{2}{m{0.95\columnwidth}}{\textbf{\texttt{BLIP-2-11B:}} \textcolor{red!80} {Yellow}.} \\

\midrule

\cellcolor{myblue!40}\textbf{Image} & \cellcolor{myblue!40}\textbf{Question \& Answer} \\ 
\midrule
\addlinespace[1mm]
\includegraphics[width=0.30\columnwidth]{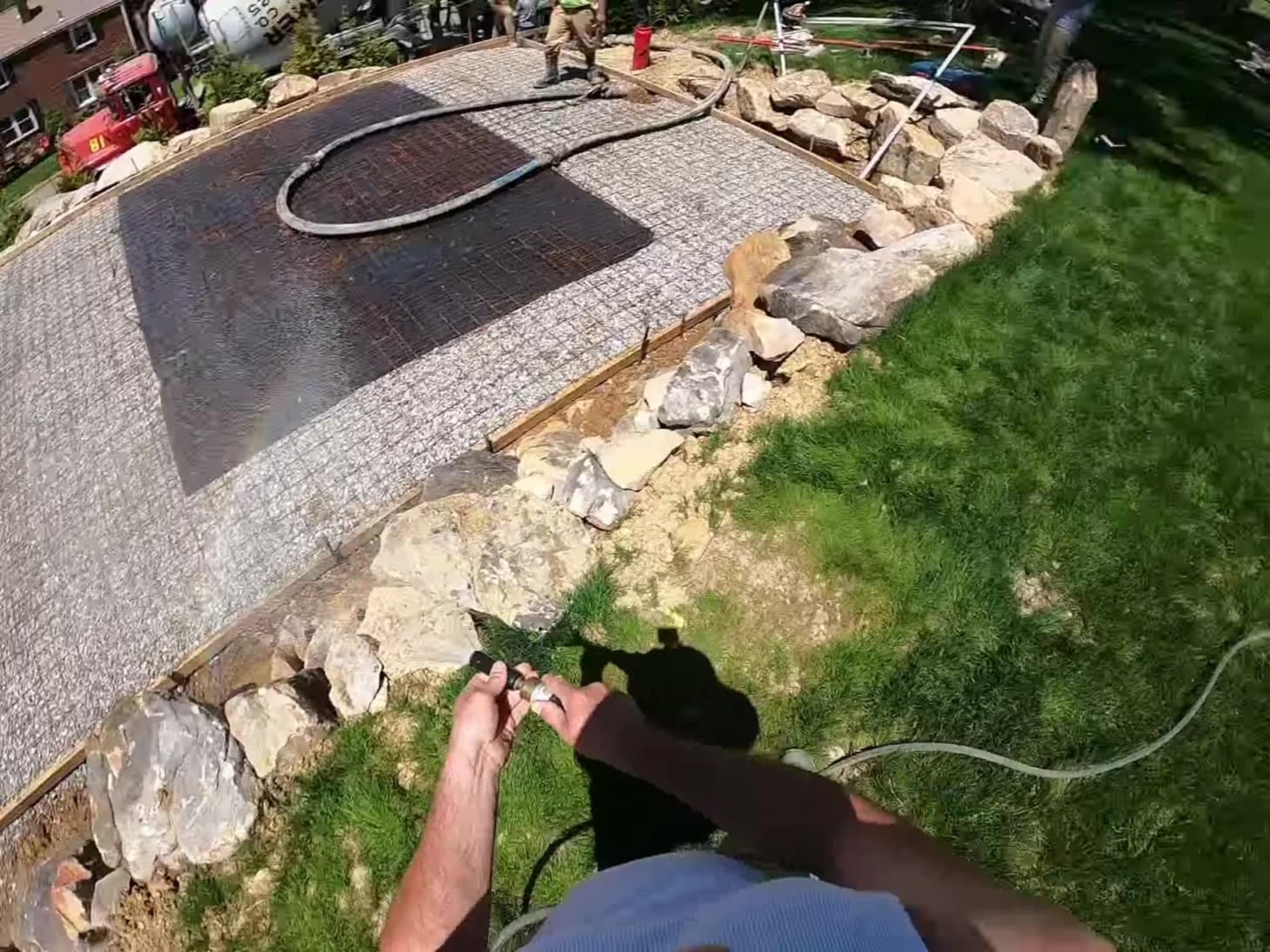} & \makecell[l]{\textit{Question:} What's the use of the thing \\ in my hands? \\ \\ \textit{Answer:} Watering.} \\
\midrule
\multicolumn{2}{l}{\cellcolor{red!20}\textbf{Model Predictions}}\\
\midrule
\addlinespace[1mm]
\multicolumn{2}{m{0.95\columnwidth}}{\textbf{\texttt{InstructBLIP-7B:}} \textcolor{red!80} {Hose nozzle}.} \\
\multicolumn{2}{m{0.95\columnwidth}}{\textbf{\texttt{BLIP-2-11B:}} \textcolor{red!80} {A hose}.} \\
\multicolumn{2}{m{0.95\columnwidth}}{\textbf{\texttt{LLaVA-13B-Vicuna:}} The thing in your hands is \textcolor{red!80} {a garden hose}.} \\

\midrule

\cellcolor{myblue!40}\textbf{Image} & \cellcolor{myblue!40}\textbf{Question \& Answer} \\ 
\midrule
\addlinespace[1mm]
\includegraphics[width=0.30\columnwidth]{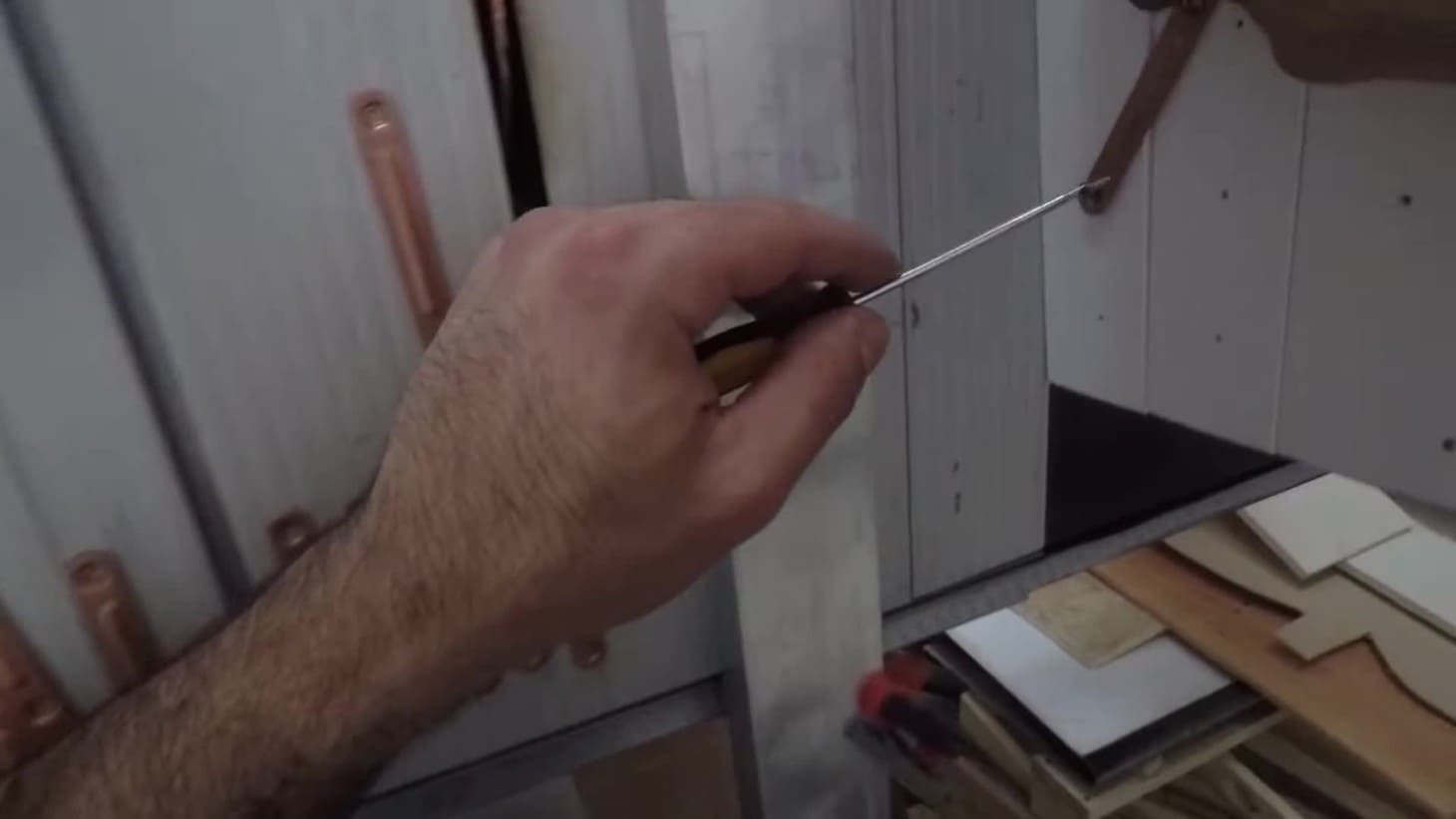} & \makecell[l]{\textit{Question:} What can the thing in my left \\ hand be used for? \\ \\ \textit{Answer:} Screwing.} \\
\midrule
\multicolumn{2}{l}{\cellcolor{red!20}\textbf{Model Predictions}}\\
\midrule
\addlinespace[1mm]
\multicolumn{2}{m{0.95\columnwidth}}{\textbf{\texttt{BLIP-2-11B:}} \textcolor{red!80}{A screwdriver.}} \\
\multicolumn{2}{m{0.95\columnwidth}}{\textbf{\texttt{InstructBLIP-11B:}} \textcolor{red!80}{Screwdriver.}} \\
\multicolumn{2}{m{0.95\columnwidth}}{\textbf{\texttt{LLaVA-7B:}} The thing in your left hand can be used as \textcolor{red!80}{A screwdriver}.} \\

\bottomrule

\end{tabular}
\end{tcolorbox}
\caption{Case studies (\textcolor{red!80}{wrong}) in attribute and affordance dimensions.}
\label{tab:exs_att_aff}
\end{figure}

\begin{figure}[t!]
\centering
\footnotesize
\begin{tcolorbox}[
  enhanced,        
  colback=white,   
  colframe=black,  
  arc=3mm,         
  width=\columnwidth, 
  boxrule=0.8pt,    
  left=5pt,  
  right=1pt  
]
\begin{tabular}{m{0.30\columnwidth}p{0.59\columnwidth}}
\toprule

\cellcolor{myblue!40}\textbf{Image} & \cellcolor{myblue!40}\textbf{Question \& Answer} \\ 
\midrule
\addlinespace[1mm]
\includegraphics[width=0.30\columnwidth]{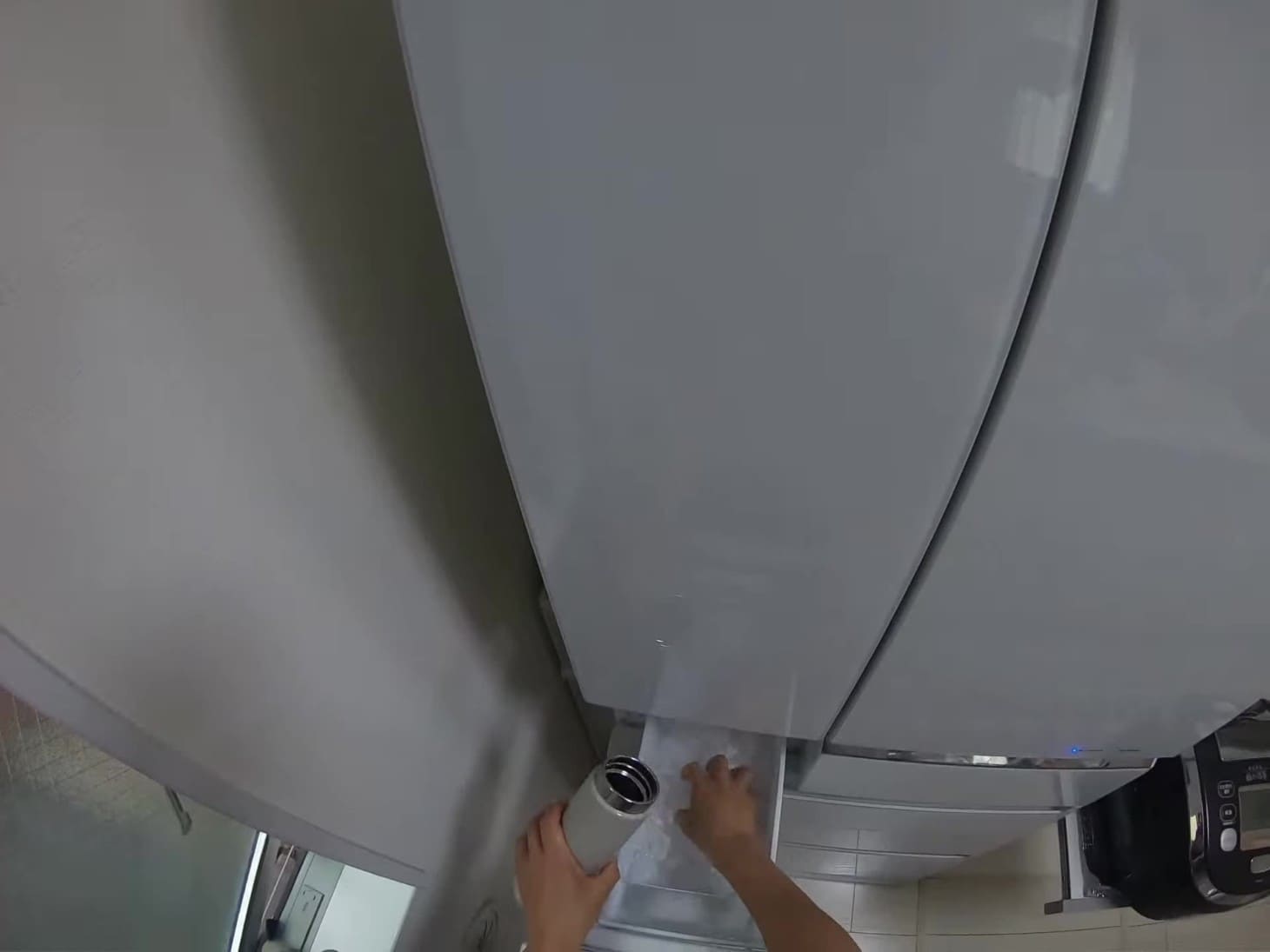} & \makecell[l]{\textit{Question:} What am I doing now?
\\ \\ \textit{Answer:} Putting ice into a bottle.} \\
\midrule
\multicolumn{2}{l}{\cellcolor{red!20}\textbf{Model Predictions}}\\
\midrule
\addlinespace[1mm]
\multicolumn{2}{m{0.95\columnwidth}}{\textbf{\texttt{GPT-4V:}} Cleaning a \textcolor{red!80} {range hood}.} \\
\multicolumn{2}{m{0.95\columnwidth}}{\textbf{\texttt{InstructBLIP-13B:}} Cleaning \textcolor{red!80}{dishwasher}.} \\
\multicolumn{2}{m{0.95\columnwidth}}{\textbf{\texttt{BLIP-2-11B:}} Installing a \textcolor{red!80} {washing machine}.} \\

\bottomrule
\end{tabular}
\end{tcolorbox}
\caption{Cases studies (\textcolor{red!80}{wrong}) in the activity dimensions.}
\label{tab:act}
\end{figure}

\noindent\textbf{Cases on object.}
In the existence dimension, GPT-4V and other open-source VLMs still have a hard time dealing with unusual cases as shown in Figure~\ref{tab:existence_gpt_2}.
In the top case, GPT-4V cannot detect the exact location of the mentioned object. As for the other two cases, VLMs even cannot identify the detailed objects.
As shown in \Cref{tab:exs_att_aff}, in the first case, the VLMs also locate the wrong place, inferring the glove rather than ``the cap of the bottle''.
In the second and third cases, VLMs only answer the name of objects rather than the specific attribute or affordance.

\noindent\textbf{Cases on activity.}
In the activity dimension, GPT-4V and other models also have the problem of not being able to correctly detect objects as shown in \Cref{tab:act}, which leads to the models answering activities that are almost completely unrelated to the answer.

\begin{figure}[t!]
\centering
\footnotesize
\begin{tcolorbox}[
  enhanced,        
  colback=white,   
  colframe=black,  
  arc=3mm,         
  width=\columnwidth, 
  boxrule=0.8pt,    
  left=5pt,  
  right=1pt  
]
\begin{tabular}{m{0.30\columnwidth}p{0.59\columnwidth}}
\toprule

\cellcolor{myblue!40}\textbf{Image} & \cellcolor{myblue!40}\textbf{Original Question \& Answer} \\ 
\midrule
\addlinespace[1mm]
\includegraphics[width=0.30\columnwidth]{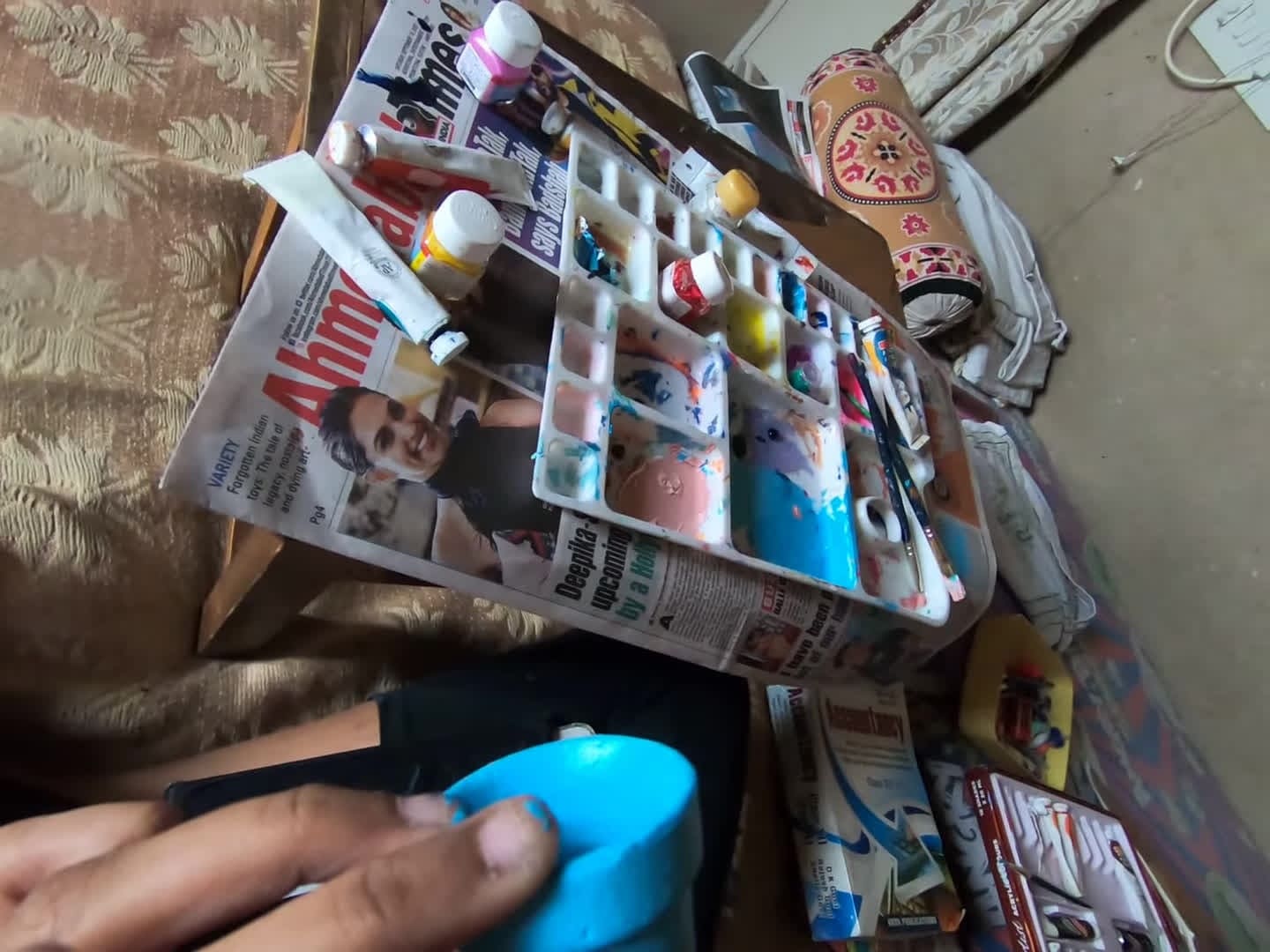} & \makecell[l]{\textit{Question:} Am I in the bedroom?
 \\ \\ \textit{Answer:} Yes.} \\
\midrule
\multicolumn{2}{l}{\cellcolor{red!20}\textbf{Model Predictions}}\\
\midrule
\addlinespace[1mm]
\multicolumn{2}{m{0.95\columnwidth}}{\textbf{\texttt{GPT-4V:}} \textcolor{red!80}{No}.} \\
\multicolumn{2}{m{0.95\columnwidth}}{\textbf{\texttt{BLIP-2-6.7B:}} \textcolor{red!80}{No}, I am in the \textcolor{red!80}{living room}.} \\
\midrule
\multicolumn{2}{l}{\cellcolor{myblue!40}\textbf{Changed Question Format \& Answer}}\\
\midrule
\addlinespace[1mm]
\multicolumn{2}{m{0.95\columnwidth}}{\textit{Question:} Where am I?} \\ 
\multicolumn{2}{m{0.95\columnwidth}}{Answer: In the bedroom.} \\
\midrule
\multicolumn{2}{l}{\cellcolor{red!20}\textbf{Model Predictions}}\\
\midrule
\addlinespace[1mm]
\multicolumn{2}{m{0.95\columnwidth}}{\textbf{\texttt{GPT-4V:}} The environment in the image appears to be \textcolor{red!80}{a living area or a space used for arts and crafts}, given the presence of newspapers, paint, and painting supplies. It does not look like a typical bedroom setup.} \\

\midrule

\cellcolor{myblue!40}\textbf{Image} & \cellcolor{myblue!40}\textbf{Question \& Answer} \\ 
\midrule
\addlinespace[1mm]
\includegraphics[width=0.30\columnwidth]{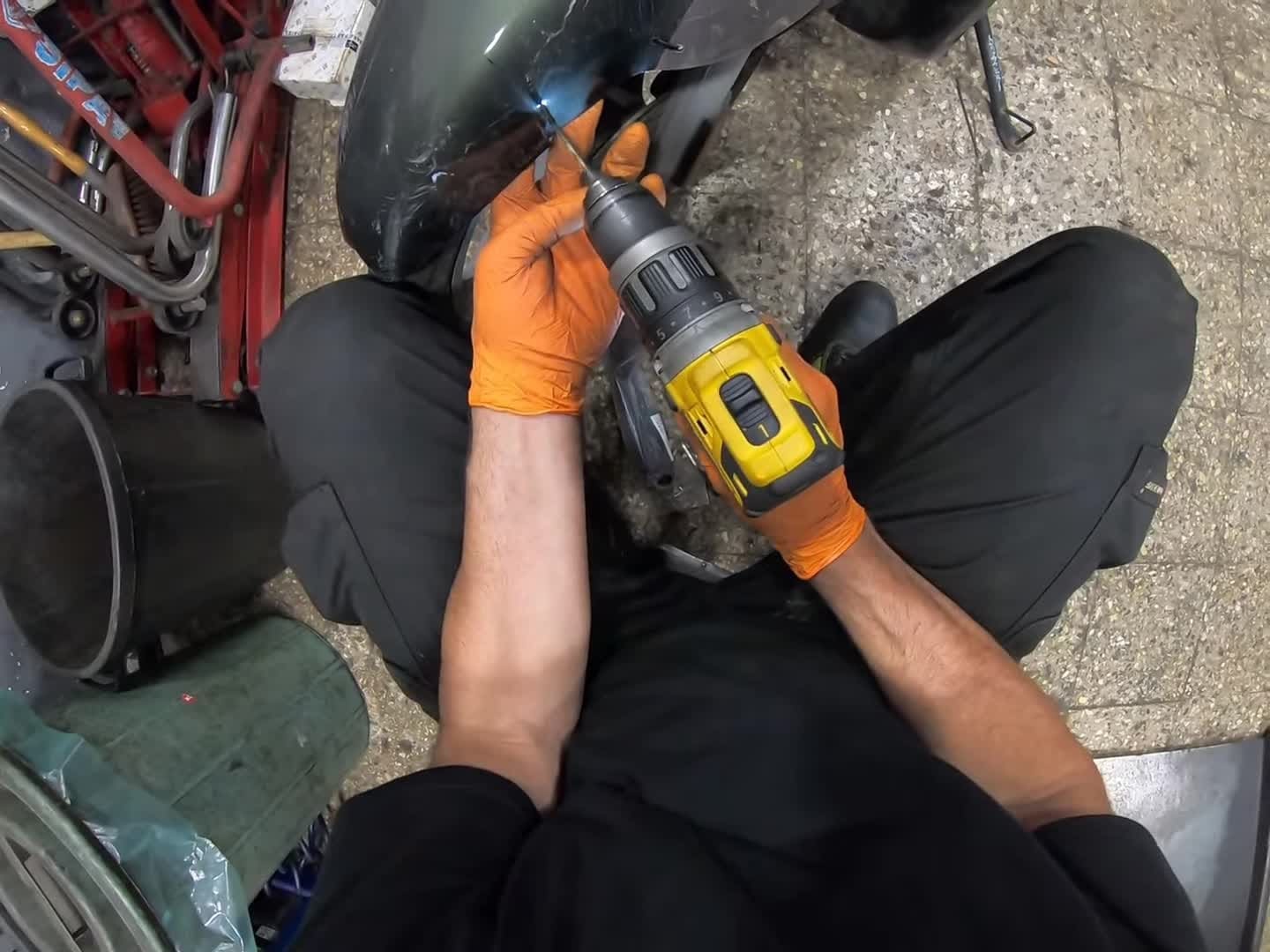} & \makecell[l]{\textit{Question:} Where are the buckets, on my \\ left or on my right or both? \\ \\ \textit{Answer:} On my left.} \\
\midrule
\multicolumn{2}{l}{\cellcolor{red!20}\textbf{Model Predictions}}\\
\midrule
\addlinespace[1mm]
\multicolumn{2}{m{0.95\columnwidth}}{\textbf{\texttt{GPT-4V:}} \textcolor{red!80}{Both}. (There are buckets on both your left and right sides. On the left, there appears to be a black bucket, and on the right, there is a green bucket.)} \\
\multicolumn{2}{m{0.95\columnwidth}}{\textbf{\texttt{BLIP-2-11B:}} On my left.} \\
\multicolumn{2}{m{0.95\columnwidth}}{\textbf{\texttt{PandaGPT-7B:}} The buckets are on my left side.} \\

\bottomrule
\end{tabular}
\end{tcolorbox}
\caption{Cases studies (\textcolor{red!80}{wrong}) in the location and spatial relationship dimensions.}
\label{tab:loc_spa}
\end{figure}

\noindent\textbf{Cases on localization.}
In the location dimension, as the first case in \Cref{tab:loc_spa}, GPT-4V and other VLMs cannot correctly detect the scene due to unexpected items. 
Even after changing the question format, GPT-4V still misunderstands the environment according to the unexpected items.
For the spatial relationship dimension, the second case in \Cref{tab:loc_spa} shows that GPT-4V is not able to recognize the egocentric view and cannot distinguish between left and right, while other VLMs can.

\begin{figure}[h!]
\centering
\footnotesize
\begin{tcolorbox}[
  enhanced,        
  colback=white,   
  colframe=black,  
  arc=3mm,         
  width=\columnwidth, 
  boxrule=0.8pt,    
  left=5pt,  
  right=1pt  
]
\begin{tabular}{m{0.30\columnwidth}p{0.59\columnwidth}}
\toprule

\cellcolor{myblue!40}\textbf{Image} & \cellcolor{myblue!40}\textbf{Question \& Answer} \\ 
\midrule
\addlinespace[1mm]
\includegraphics[width=0.30\columnwidth]{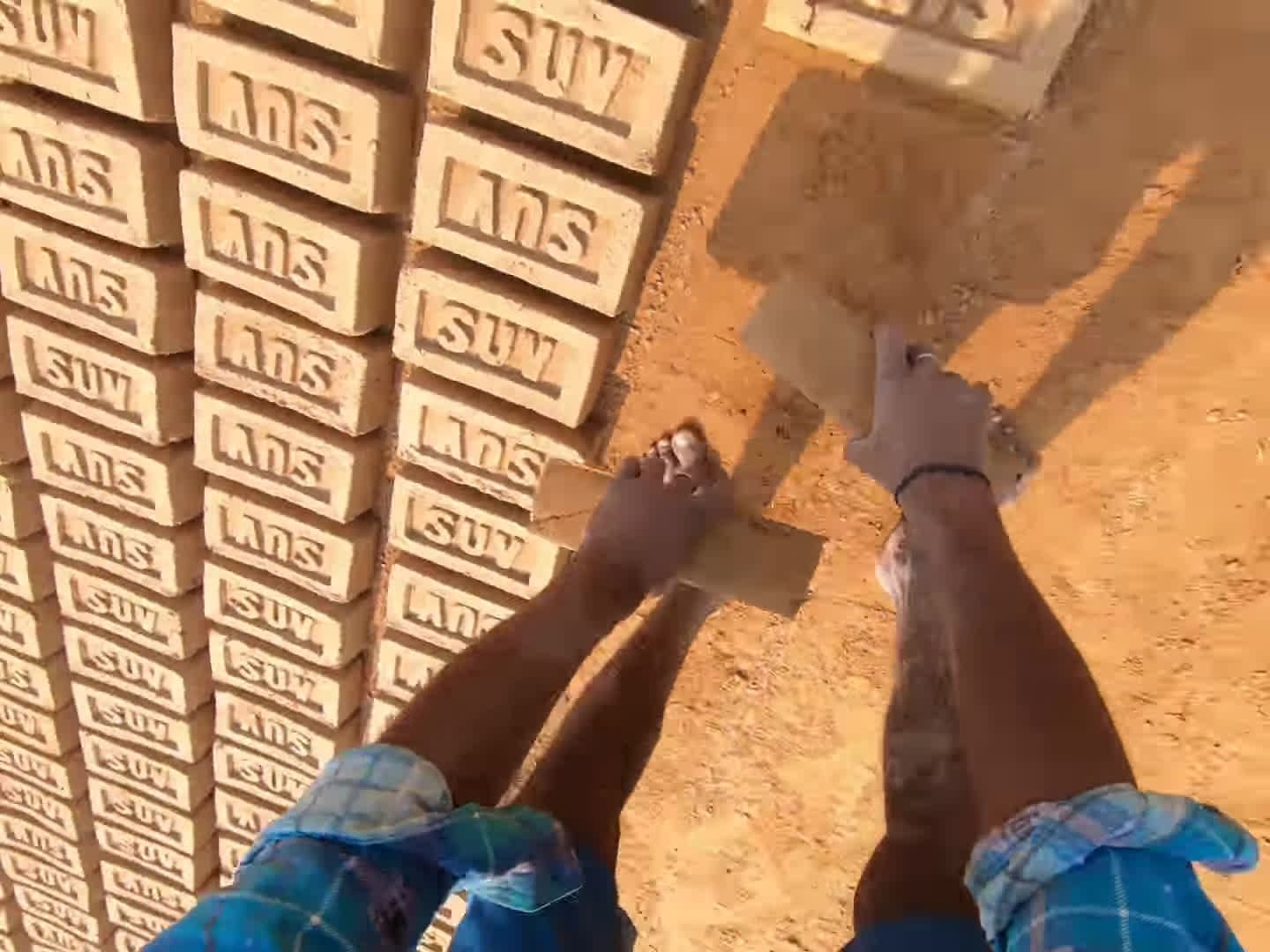} & \makecell[l]{\textit{Question:} How many bricks am I holding? \\ \\ \textit{Answer:} Two.} \\
\midrule

\multicolumn{2}{l}{\cellcolor{red!20}\textbf{Model Predictions}}\\
\midrule
\addlinespace[1mm]

\multicolumn{2}{m{0.95\columnwidth}}{\textbf{\texttt{BLIP-2-6.7B:}} \textcolor{red!80}{1,000,000,000,000,000,000,000,000,000,000,000,} \textcolor{red!80}{000,000,000}} \\
\multicolumn{2}{m{0.95\columnwidth}}{\textbf{\texttt{BLIP-2-11B:}} \textcolor{red!80} {A lot of} bricks.} \\
\multicolumn{2}{m{0.95\columnwidth}}{\textbf{\texttt{PandaGPT-7B:}} I am holding two bricks.} \\
\multicolumn{2}{m{0.95\columnwidth}}{\textbf{\texttt{PandaGPT-13B:}} I am holding two bricks.} \\

\bottomrule
\end{tabular}
\end{tcolorbox}
\caption{Cases studies (\textcolor{red!80}{wrong}) in the counting dimensions.}
\label{tab:cou}
\end{figure}

\noindent\textbf{Cases on reasoning.}
In the counting dimension, as shown in~\cref{tab:cou}, we find that some VLMs can not distinguish the specific location reference, such as ``I holding''. Moreover, when the number is large, it cannot say the exact amount.

\begin{figure}[t!]
\centering
\footnotesize
\begin{tcolorbox}[
  enhanced,        
  colback=white,   
  colframe=black,  
  arc=3mm,         
  width=\columnwidth, 
  boxrule=0.8pt,    
  left=5pt,  
  right=1pt  
]
\begin{tabular}{m{0.30\columnwidth}m{0.59\columnwidth}}
\toprule
\cellcolor{myblue!40}\textbf{Image} & \cellcolor{myblue!40}\textbf{Question \& Answer} \\ 
\midrule
\addlinespace[1mm]
\includegraphics[width=0.30\columnwidth]{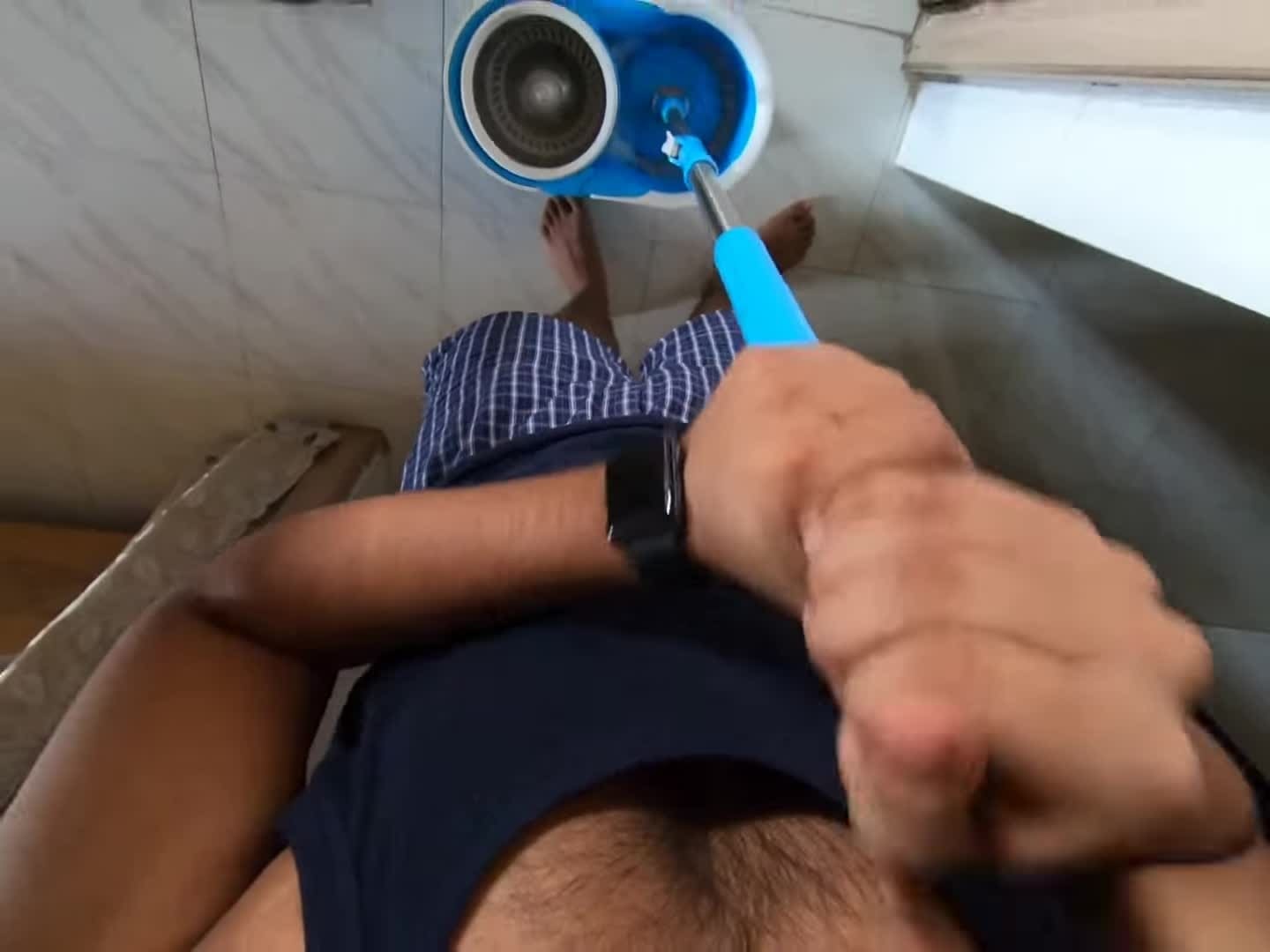} & \makecell[l]{\textit{Question:} What will I do next? \\ \textit{Answer:} Mop the floor.} \\
\midrule
\multicolumn{2}{l}{\cellcolor{red!20}\textbf{Model Predictions}}\\
\midrule
\addlinespace[1mm]

\multicolumn{2}{m{0.95\columnwidth}}{\textbf{\texttt{BLIP-2-11B:}} I will use the \textcolor{red!80} {vacuum} to clean the floor.} \\
\multicolumn{2}{m{0.95\columnwidth}}{\textbf{\texttt{InstructBLIP-13B:}} \textcolor{red!80} {Brush teeth}.} \\
\multicolumn{2}{m{0.95\columnwidth}}{\textbf{\texttt{PandaGPT-13B:}} I will \textcolor{red!80} {brush my teeth}.} \\

\midrule
\addlinespace[1mm]
\cellcolor{myblue!40}\textbf{Image} & \cellcolor{myblue!40}\textbf{Question \& Answer} \\ 
\midrule
\addlinespace[1mm]
\includegraphics[width=0.30\columnwidth]{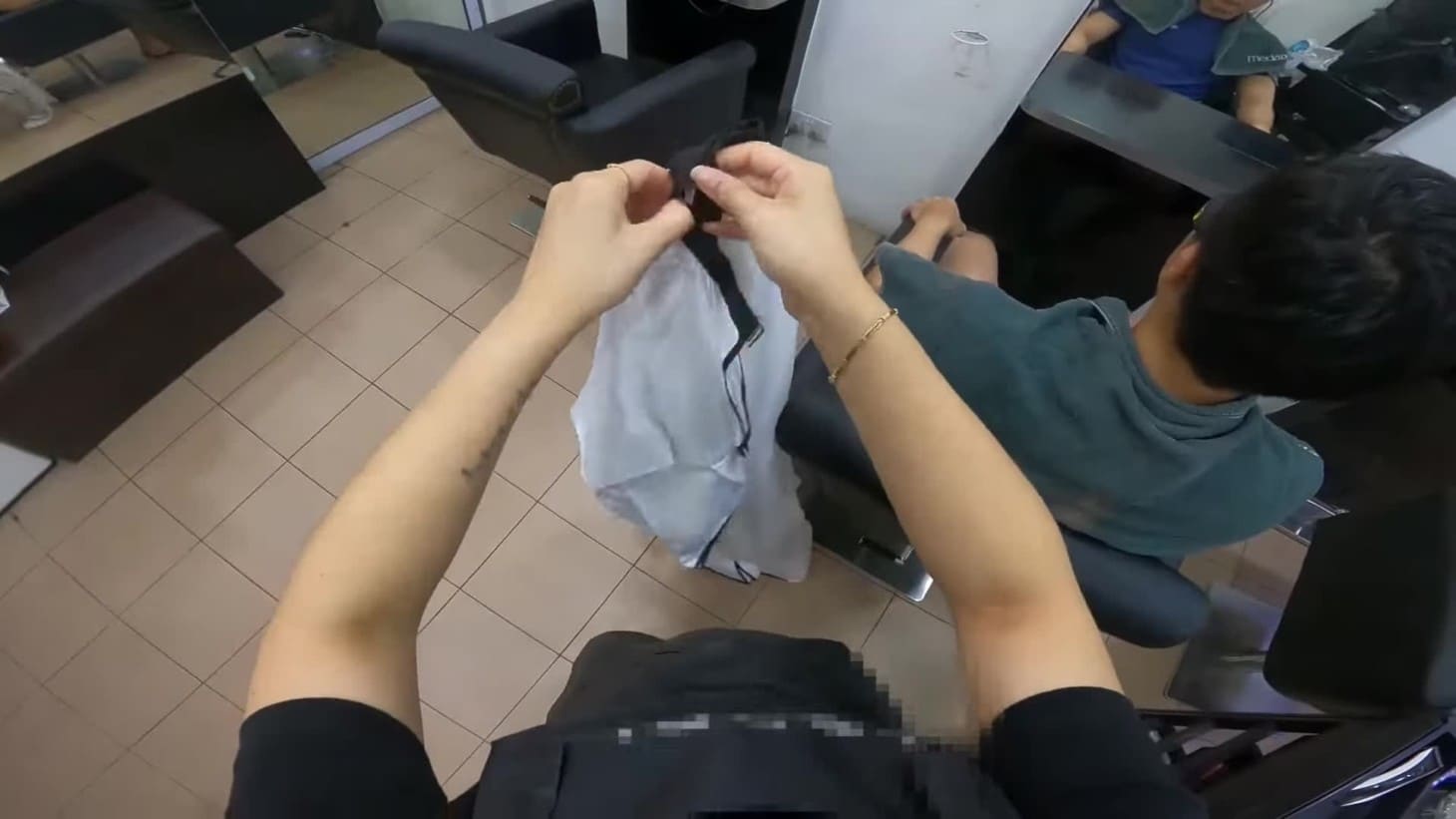} & \makecell[l]{\textit{Question:} What will I do next? \\ \textit{Answer:} Haircut.} \\
\midrule
\multicolumn{2}{l}{\cellcolor{red!20}\textbf{Model Predictions}}\\
\midrule
\addlinespace[1mm]
\multicolumn{2}{m{0.95\columnwidth}}{\textbf{\texttt{LLaMA-Adapter-7B:}} I will be \textcolor{red!80} {getting a haircut}.} \\
\multicolumn{2}{m{0.95\columnwidth}}{\textbf{\texttt{BLIP-2-11B:}} \textcolor{red!80} {Get my hair cut}.} \\
\multicolumn{2}{m{0.95\columnwidth}}{\textbf{\texttt{mPLUG-owl-7B:}} I will \textcolor{red!80} {tie my hair back and get ready for my haircut}.} \\

\bottomrule
\end{tabular}
\end{tcolorbox}
\caption{Case studies (\textcolor{red!80}{wrong}) in the forecasting dimension.}
\label{tab:for_nav_ass}
\end{figure}

\noindent\textbf{Cases on forecasting.}
The first case in \Cref{tab:for_nav_ass} demonstrates that the models cannot identify the objects accurately. In the second case, the models are not able to recognize the egocentric view.

\begin{figure}[h!]
\centering
\footnotesize
\begin{tcolorbox}[
  enhanced,        
  colback=white,   
  colframe=black,  
  arc=3mm,         
  width=\columnwidth, 
  boxrule=0.8pt,    
  left=5pt,  
  right=1pt  
]
\begin{tabular}{m{0.30\columnwidth}p{0.59\columnwidth}}
\toprule

\cellcolor{myblue!40}\textbf{Image} & \cellcolor{myblue!40}\textbf{Question \& Answer} \\ 
\midrule
\addlinespace[1mm]
\includegraphics[width=0.30\columnwidth]{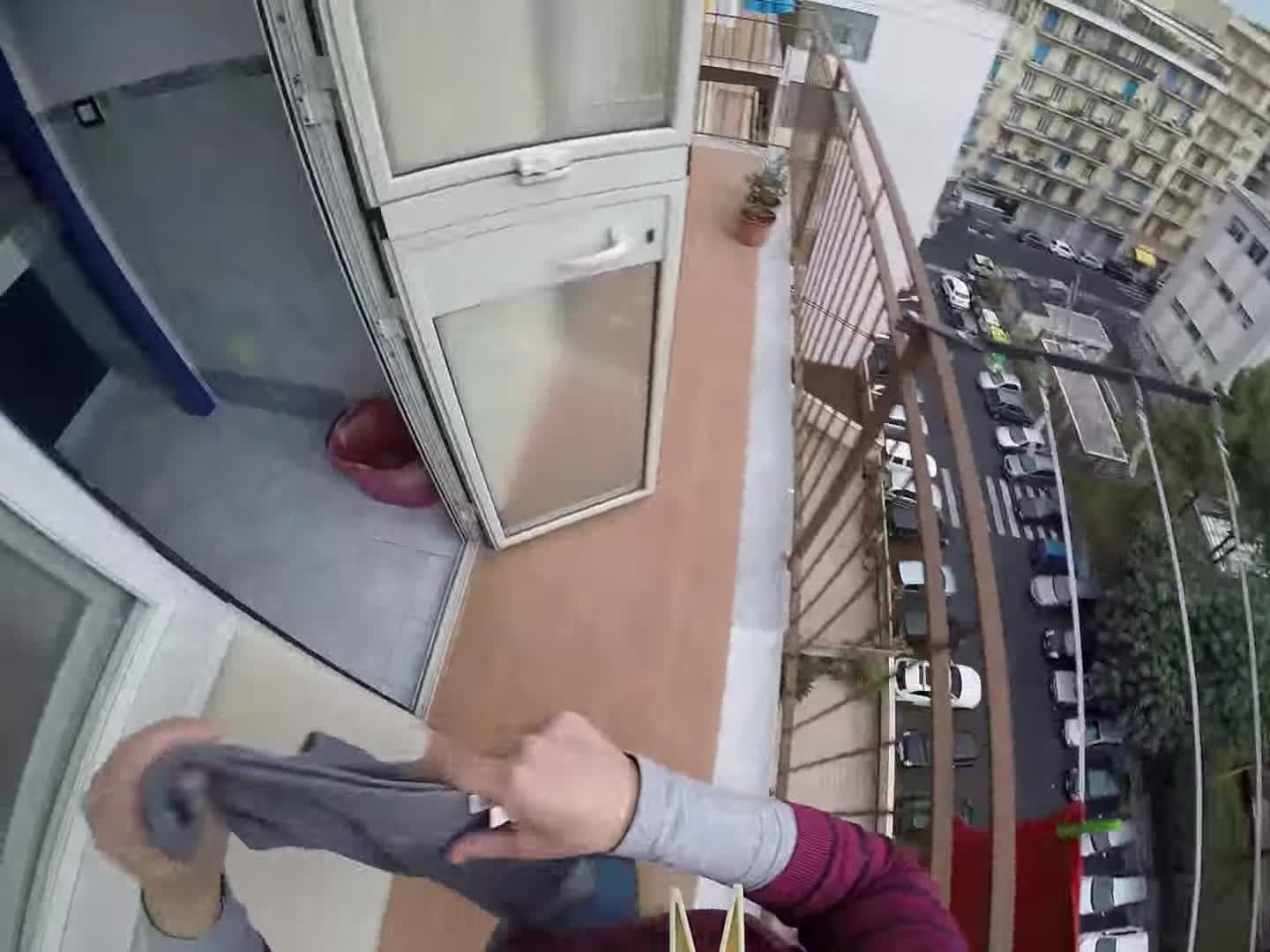} & \makecell[l]{\textit{Question:} How can I get into the room? \\ \\ \textit{Answer:} Take one or two steps forward,  \\ then turn left and continue walking.} \\
\midrule
\multicolumn{2}{l}{\cellcolor{red!20}\textbf{Model Predictions}}\\
\midrule
\addlinespace[1mm]
\multicolumn{2}{m{0.95\columnwidth}}{\textbf{\texttt{InstructBLIP-13B:}} \textcolor{red!80} {Elevator}.} \\
\multicolumn{2}{m{0.95\columnwidth}}{\textbf{\texttt{mPLUG-owl-7B:}} You can use the \textcolor{red!80} {ladder} to climb up to the balcony and enter the room.} \\
\multicolumn{2}{m{0.95\columnwidth}}{\textbf{\texttt{LLaMA-Adapter-7B:}} You can enter the room by using the \textcolor{red!80} {ladder} that is already set up outside the window.} \\

\midrule

\cellcolor{myblue!40}\textbf{Image} & \cellcolor{myblue!40}\textbf{Question \& Answer} \\ 
\midrule
\addlinespace[1mm]
\includegraphics[width=0.30\columnwidth]{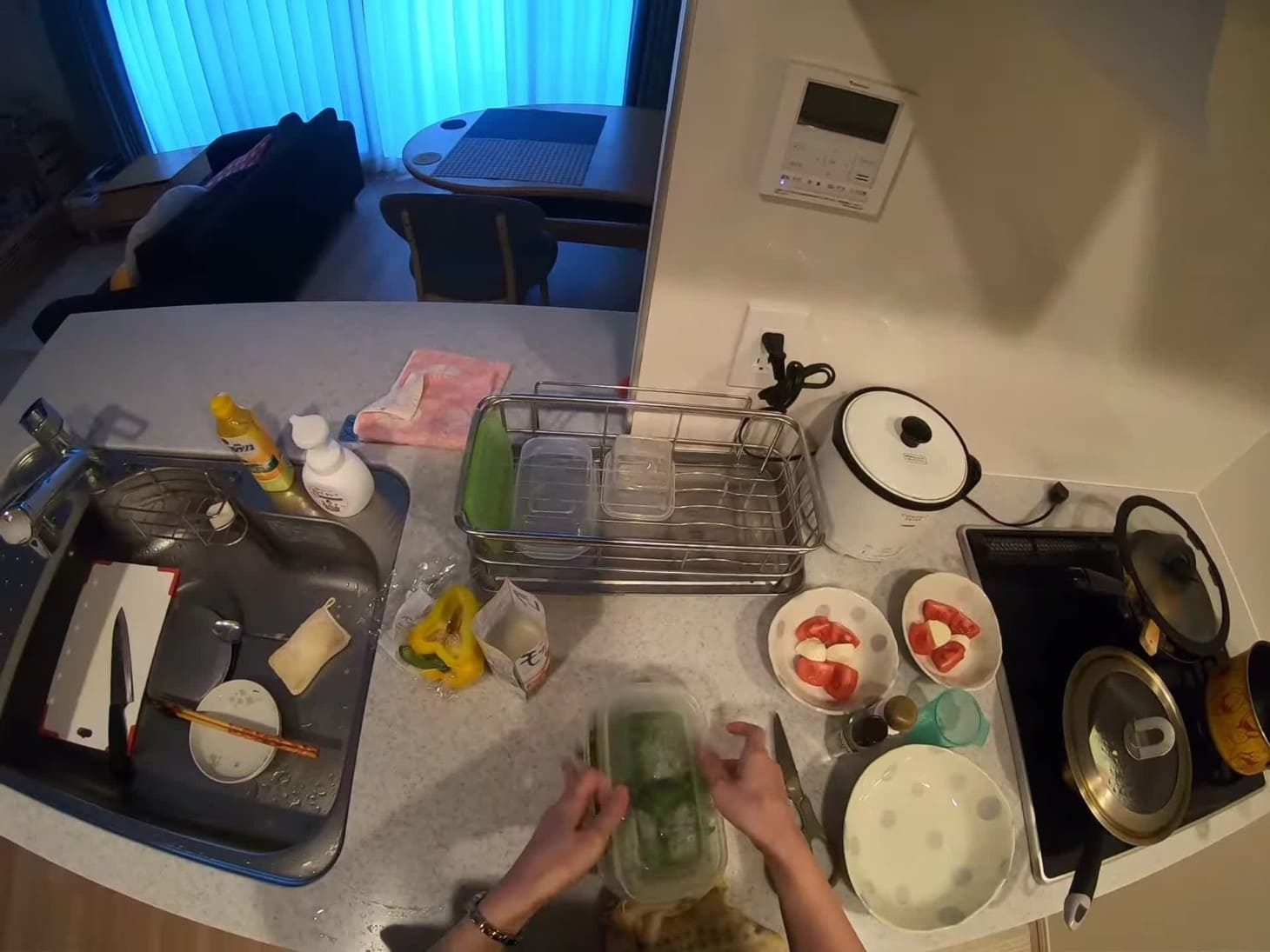} & \makecell[l]{\textit{Question:} How do I pour the contents of \\ the container in my hand into the plate on \\ my right? \\ \\ \textit{Answer:} Open the lid of the container, \\ locate the plate and pour the contents into \\ the plate.} \\
\midrule
\multicolumn{2}{l}{\cellcolor{red!20}\textbf{Model Predictions}}\\
\midrule
\addlinespace[1mm]
\multicolumn{2}{m{0.95\columnwidth}}{\textbf{\texttt{LLaVA-7B:}} To pour the contents of the container in your hand into the plate on your right, you can use a spoon to scoop up the food from the container and then carefully pour it into the plate. Make sure to hold the spoon at an angle to ensure that the food flows smoothly into the plate without spilling.} \\
\multicolumn{2}{m{0.95\columnwidth}}{\textbf{\texttt{Otter-I-7B:}} To pour the contents of the container into the plate, follow these steps: 1. Hold the container in your hand and position it over the plate on the right. 2. Carefully pour the contents from the container into the plate, ensuring that the contents are evenly distributed across the plate. 3. Once the contents are poured into the plate, you can place the container aside.} \\

\bottomrule
\end{tabular}
\end{tcolorbox}
\caption{Case studies (\textcolor{red!80}{wrong}) in the navigation and assistance dimensions.}
\label{tab:nav_ass}
\end{figure}

\noindent\textbf{Cases on planning.}
In the navigation and assistance dimensions, models can neglect important information in the images and the answer might be too brief without details or too lengthy without getting to the point as shown in the two cases in the \Cref{tab:nav_ass}.


\section{The information of image encoder}
In the main paper, additional image information can assist the model in detecting objects on certain images, as shown in Figure~\ref{tab:som}.
However, in order to quantitatively analyze the function of this module, we conduct experiments on both extensive and attribute dimensions. The results are shown in Table~\ref{tab:som_results} where the correct answer rate of samples with SoM has decreased. We consider that the additional image marks and masks obscure the information in the original image (such as colors, object borders, etc.), resulting in incorrect model judgments. How to provide additional information without losing the original image information may be a future research direction that can be considered.

\label{app:som}
\begin{table}[htbp!]
    \centering
    \small
    \begin{tabular}{l|cc}
    \toprule
        {\textbf{Methods}} & {\textbf{Existence}} & {\textbf{Attribute}} \\
        \toprule 
        GPT-4V & 62.0 & 82.0 \\
        GPT-4V w/ SoM & 36.0 & 62.0 \\
    \bottomrule
    \end{tabular}
    \caption{Grading scores on the Existence and Attribute dimensions of GPT-4V with SoM vision encoder, evaluated by GPT-4.}
    \label{tab:som_results}
\end{table}

\section{Human Annotation}
\label{app:human_annotation}

\subsection{Annotation System}
\begin{figure}[htbp!]
    \centering
    \includegraphics[width=\linewidth]{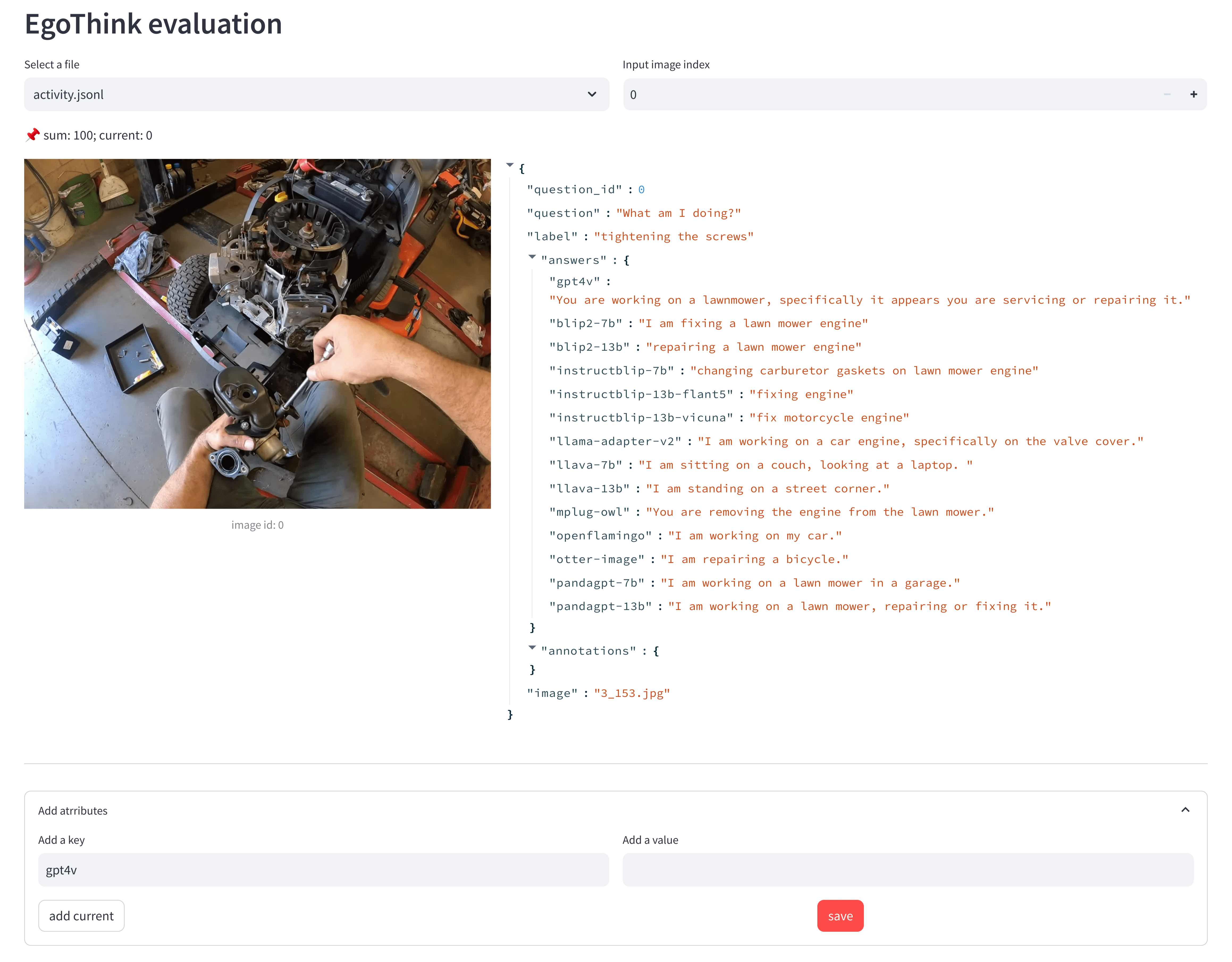}
    \caption{Our EgoThink evaluation system for manual annotations.}
    \label{fig:enter-label}
\end{figure}

In order to save human labor, we construct an annotation system based on Streamlit. Our annotation system is designed as a multi-user image and text annotation system, which can display images and provide an interactive interface for users to annotate efficiently as shown in Figure~\ref{fig:enter-label}.

\subsection{Annotation Guideline}
Here we present the detailed annotation guidelines for annotators:
1) Accuracy. The model output should be factually correct, without violating commonsense and the knowledge provided in the data. 
2) Completeness. It is acceptable that the format of the answer given by the model is different from the reference answer, but the model output should provide the key information of the reference answer or reasonable answer beyond the reference answer.
3) Logic. The answer should be logical. It should provide answers with reasonable logical sequence.
4) Language and grammar. The output should use correct spelling, vocabulary, punctuation, and grammar.

\section{Agreement}
\label{app:agreement}
\begin{figure}[htbp!]
    \centering
    \includegraphics[width=\linewidth]{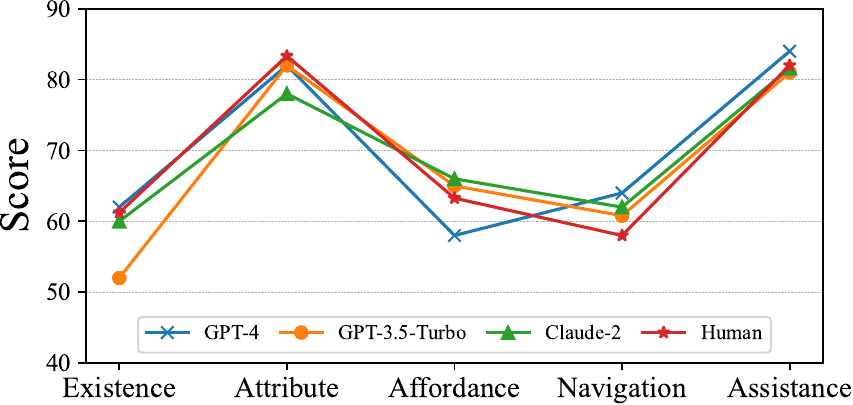}
    \caption{Average scores of GPT-4V on Object and Planning given by different evaluators.}
    \label{fig:evaluation_agreement}
\end{figure}

We select object and planning dimensions to compare the differences between evaluation models. Scores of different models evaluated by GPT-3.5-Turbo, Claude-2, GPT-4V, and human annotators are shown in \Cref{tab:gpt3.5_claude_scores,tab:human_score}, and the Pearson correlation coefficients between them are shown in Table~\ref{tab:evaluation_corr}.
As our main evaluation model, GPT-4V is scored by different evaluators (including humans), and the average scores are shown in Figure~\ref{fig:evaluation_agreement}.
In general, the consistency among GPT-4, Claude-2, and Humans is high.

\begin{table}[t!]
    \centering
    \small
    \begin{tabular}{c|ccc}
        \toprule
         & \textbf{GPT-4} & \textbf{GPT-3.5-Turbo} & \textbf{Claude-2} \\
         \midrule
        \textbf{GPT-3.5-Turbo} & 52.4 & - & -\\
        \textbf{Claude-2} & 80.0 & 53.6 & - \\
        \textbf{Human} & 68.2  & 43.6 & 68.4 \\
        \bottomrule
    \end{tabular}
    \caption{Pearson correlation coefficients between GPT-4, GPT-3.5-Turbo, Claude-2 and human evaluations on Object and Planning dimensions. }
    \label{tab:evaluation_corr}
\end{table}

\begin{table}[htbp!]
    \centering
    \small
    \begin{tabular}{l|ccc|cc}
    \toprule
        \multirow{2}{*}{\textbf{Grading methods}} & \multicolumn{3}{c|}{\textbf{Object}} & \multicolumn{2}{c}{\textbf{Planning}} \\
         & {Exist} & {Attr} & {Afford } & {Nav} & {Assist} \\
        \toprule 
       Human  & 61.2 & 83.3 & 63.3 & 58.0 & 82.0 \\
    \bottomrule
    \end{tabular}
    \caption{Grading scores for the Object and Planning dimensions for GPT-4V by human annotators.}
    \label{tab:human_score}
\end{table}

\begin{table*}[t!]
    \resizebox{\linewidth}{!}
    {
        \centering
        \small
        \begin{tabular}{l|cccccc|cccc}
            \toprule
            \multirow{3}{*}{\textbf{Methods}} & \multicolumn{6}{c|}{\textbf{Object}} & \multicolumn{4}{c}{\textbf{Planning}} \\
            \cmidrule{2-7}
            \cmidrule{8-11}
            & \multicolumn{2}{c}{Existence} & \multicolumn{2}{c}{Attribute} & \multicolumn{2}{c|}{Affordance} & \multicolumn{2}{c}{Navigation} & \multicolumn{2}{c}{Assistance} \\
            & GPT-3.5 & Claude-2 & GPT-3.5 & Claude-2 & GPT-3.5 & Claude-2 & GPT-3.5 & Claude-2 & GPT-3.5 & Claude-2\\
            \toprule
            \rowcolor[gray]{0.95} \multicolumn{11}{c}{\textbf{API-based model}} \\
            \midrule

            GPT-4V & 52.0 & 60.0 & 82.0 & 78.0 & 65.0 & 66.0  & 60.8 & 62.0 & 81.0 & 81.6\\
            
            \midrule
            \rowcolor[gray]{0.95} \multicolumn{11}{c}{\textbf{$\sim$7B Models}} \\
            \midrule
            
            OpenFlamingo  & 46.0 & 53.0 & 50.0 & 56.0 & 47.0 & 54.0 & 9.0  & 18.4 & 7.1 & 18.06  \\
            BLIP-2        & 42.0 & 50.0 & 26.0 & 37.5 & 36.7 & 46.8 & 6.0  & 8.5  & 2.2 & 10.2 \\
            Otter         & 55.0  & 52.0  &  51.0  &  56.0  & 40.6  &  48.0  &  34.0  &  34.3  & 53.3 &  51.1 \\
            PandaGPT      & 50.0  & 52.0  &  57.0  &  58.0  &  43.0  &  53.0  &  36.0  &  38.8  &  70.0 &  60.4 \\
            InstructBLIP  & 46.0  & 52.0  &  26.0  &  36.0  &  40.0  &  55.0  &  11.0  &  28.0  &  23.0 &  44.0 \\
            LLaMA-Adapter & 48.0  & 49.0  &  59.0  &  61.0  &  54.0  &  47.0  &  40.0  &  49.0  &  65.8 &  67.0 \\
            MiniGPT-4     & 61.0  & 61.0  &  58.0  &  58.0  &  35.7  &  55.0  &  29.0  &  20.8  &  57.0 &  57.0 \\
            mPLUG-owl     & 56.0  & 63.0  &  56.0  &  60.0  &  51.0  &  57.0  &  38.0  &  39.0  &  55.8 &  58.8 \\
            LLaVA         & 63.0  & 69.0  &  59.0  &  60.0  &  38.0  &  53.0  &  40.0  &  38.0  &  67.0 &  66.0 \\
            LLaVA 1.5     & 35.0  & 33.7  &  43.0  &  50.0  &  37.8  &  63.3  &  39.0  &  33.7  &  73.0 &  66.7 \\
            
            \midrule 
            \rowcolor[gray]{0.95} \multicolumn{11}{c}{\textbf{$\sim$13B Models}} \\
            \midrule

            PandaGPT        & 51.0           & 57.0  &  53.0  &  55.0  &  49.0  &  48.0  &  46.0  &  45.0  &  81.0 &  79.2 \\
            InstructBLIP(V) & 51.0           & 54.0  &  49.0  &  53.0  &  51.0  &  55.0  &  10.0  &  23.0  &  19.0 &  39.0 \\
            BLIP-2          & 49.0           & 51.0  &  58.0  &  61.0  &  45.0  &  54.0  &  19.0  &  24.0  &  25.0 &  33.0 \\
            LLaVA           & 66.0           & 68.0  &  62.0  &  64.0  &  52.0  &  62.0  &  38.0  &  36.0  &  67.0 &  73.0 \\
            InstructBLIP(F) & 63.0           & 66.0  &  57.0  &  68.0  &  45.0  &  52.0  &  10.0  &  23.0  &  11.0 &  31.0 \\
            LLaVA 1.5       & 67.0           & 70.0  &  54.0  &  57.0  &  48.0  &  58.0  &  33.0  &  45.0  &  47.0 &  56.0 \\
            LLaVA (L2)      & 68.0  & 74.0  &  64.0  &  65.0  &  36.2  &  49.0  &  48.0  &  43.0  &  78.6 &  74.6 \\
            \bottomrule
        \end{tabular}
    }
    \caption{Grading scores for the Object and Planning dimensions of various VLMs evaluated by GPT-3.5-turbo and Claude-2. Note that GPT-3.5-Turbo and Claude-2 may not exactly give 0, 0.5, and 1 scores or successfully give a score, so the effective number of samples may be less than 50 or 100.}
    \label{tab:gpt3.5_claude_scores}
\end{table*}